%% file: main.tex
\documentclass{article}

\usepackage{microtype}
\usepackage{graphicx}
\usepackage{subfigure}
\usepackage{booktabs} %

\usepackage{hyperref}

\usepackage{xcolor}

\usepackage[accepted]{icml2020}

\usepackage{enumitem}

\usepackage{amsfonts}       %
\usepackage{amssymb}
\usepackage{nicefrac}       %

\newcommand{\bm}{\mathbf{m}}
\newcommand{\bw}{\mathbf{w}}

\newcommand{\bzero}{\boldsymbol{0}}
\newcommand{\bJ}{\mathbf{J}}
\newcommand{\bx}{\mathbf{x}}
\newcommand{\bh}{\mathbf{h}}

\newcommand{\bell}{\boldsymbol{\ell}}

\newcommand{\bbeta}{\boldsymbol{\beta}}
\newcommand{\bor}{\tiny{BOD}}
\newcommand{\eor}{\tiny{EOD}}
\newcommand{\mbor}{\mbox{\bor}}
\newcommand{\meor}{\mbox{\eor}}
\newcommand{\te}{\text{e}}

\newcommand{\Jrec}{\bJ^\text{rec}}
\newcommand{\Jinp}{\bJ^\text{inp}}
\newcommand{\evalat}[1]{\rvert_{#1}}

\definecolor{blue}{HTML}{2b6cb0}
\newcommand{\emphasis}[1]{\textcolor{blue}{\textbf{#1}}}

\definecolor{orange}{HTML}{c05621}
\newcommand{\negation}[1]{\textcolor{orange}{\textbf{#1}}}

\definecolor{teal}{HTML}{285E61}

\newcommand{\hmod}{\bh^{\text{mod}}}

\usepackage{amsmath} %
\usepackage{soul}

\icmltitlerunning{How recurrent networks implement contextual processing}

\begin{document}

\twocolumn[
\icmltitle{How recurrent networks implement contextual processing in sentiment analysis}

\icmlsetsymbol{equal}{*}

\begin{icmlauthorlist}
\icmlauthor{Niru Maheswaranathan}{equal,goo}
\icmlauthor{David Sussillo}{equal,goo}
\end{icmlauthorlist}

\icmlaffiliation{goo}{Google Research, Brain Team, Mountain View, California, USA}

\icmlcorrespondingauthor{Niru Maheswaranathan}{nirum@google.com}
\icmlcorrespondingauthor{David Sussillo}{sussillo@google.com}

\icmlkeywords{Recurrent Neural Networks, Interpretability, Sentiment Classification, Natural Language Processing, NLP}

\vskip 0.3in
]

\printAffiliationsAndNotice{\icmlEqualContribution}

\input{text/00-abstract.tex}
\input{text/01-introduction.tex}
\input{text/02-preliminaries.tex}

\input{text/03-synthetic.tex}
\input{text/04-results.tex}

\input{text/05-related-work.tex}
\input{text/06-discussion.tex}

\input{text/07-acknowledgements.tex}  %

\bibliography{refs}
\bibliographystyle{icml2020}

\pagebreak
\newpage
\appendix
\onecolumn

\input{text/08-supplemental.tex}

\end{document}

%% file: text/00-abstract.tex
\begin{abstract}
Neural networks have a remarkable capacity for contextual processing\textemdash{}using recent or nearby inputs to modify processing of current input. For example, in natural language, contextual processing is necessary to correctly interpret negation (e.g. phrases such as ``not bad''). However, our ability to understand \textit{how} networks process context is limited. Here, we propose general methods for reverse engineering recurrent neural networks (RNNs) to identify and elucidate contextual processing. We apply these methods to understand RNNs trained on sentiment classification.
This analysis reveals inputs that induce contextual effects, quantifies the strength and timescale of these effects, and identifies sets of these inputs with similar properties. Additionally, we analyze contextual effects related to differential processing of the beginning and end of documents. Using the insights learned from the RNNs we improve baseline Bag-of-Words models with simple extensions that incorporate contextual modification, recovering greater than 90\% of the RNN's performance increase over the baseline. This work yields a new understanding of how RNNs process contextual information, and provides tools that should provide similar insight more broadly.
\end{abstract}

%% file: text/01-introduction.tex
\section{Introduction}
Neural networks do a remarkable job at learning structure in natural data.
These architectures exploit complex relationships between inputs to perform tasks at state-of-the-art levels~\cite{lecun2015deep}.
Despite this amazing performance, we still only have a rudimentary understanding of exactly how these networks work~\cite{castelvecchi2016can}.

Rigorously understanding how networks solve important tasks is a central challenge in deep learning. This understanding is lacking because we use complex optimization procedures to set the parameters of ever more complex network architectures. Our inability to fundamentally understand how a trained system works makes it difficult to identify biases in the network or training data, control for adversarial input, bracket or bound network behavior, suggest ways of improving efficiency or accuracy, and elucidate the (potentially simple) core mechanisms underlying the task. 

In this work, we focus on building general tools and analyses to understand how recurrent neural networks (RNNs) process contextual effects.
By \textit{contextual effects}, we mean effects where the interpretation of an input depends on surrounding inputs.
For example, in natural language, the interpretation of words is modified by preceding words for negation (``not bad'' vs ``bad'') or emphasis (``extremely awesome'' vs ``awesome'')~\cite{quirk1985,horn2000negation}.
We analyze networks trained to perform sentiment classification, a commonly studied natural language processing (NLP) task~\cite{Wiegand2010,mohammad2017challenges,Zhang2018}. %

When interpreting or understanding how recurrent networks work, one line of research uses attribution methods to analyze sensitivity to particular inputs~\cite{simonyan2013deep,lime,li2015visualizing,arras2017explaining,Murdoch2018}. These methods utilize gradients of the network loss or unit activations with respect to inputs. A second line of research uses tools from dynamical systems analysis to decompose RNN trajectories~\cite{zipser1991recurrent,tsung1995phase,casey1996dynamics,rodriguez1999recurrent,Sussillo2013,Mante2013,cueva2018emergence,jordan2019gated,maheswaranathan2019b}. %
Here, we merge these lines of research to understand contextual processing of large numbers of inputs in recurrent systems.

In particular, our key insight is that rich contextual processing can be understood as a composition of precisely organized input-driven state deflections combined with highly organized transient dynamics.  Critically, the former can be as understood using Jacobians with respect to the input, analogous to attribution methods, while the latter can be understood through Jacobians with respect to the recurrent state, analogous to work on RNN dynamics.

\begin{figure*}[ht]
    \begin{center}
    \centerline{\includegraphics[width=0.75\textwidth]{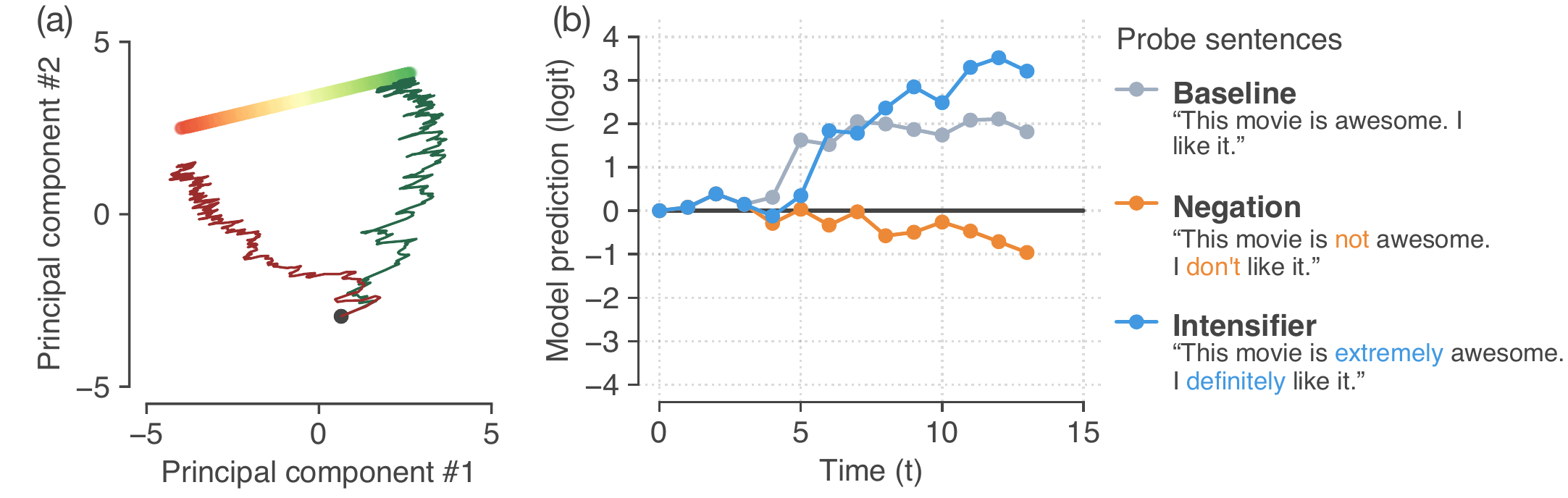}}
    \caption{RNNs and contextual processing. (a) Approximate line attractor dynamics in a GRU trained to perform sentiment classification. PCA visualization of trajectories of the system during an example positive (green) and negative (red) review. The approximate line attractor is visualized as the light red to green line, colored according to the readout (logit). (b) Evolution of RNN predictions (logits) when processing example reviews over time (tokens in the input sentence). Gray (baseline): ``This movie is awesome. I like it'' (with pad tokens inserted to align it with the other examples), Blue (intensifier): ``This movie is extremely awesome. I definitely like it.'' Orange (negation): ``This movie is not awesome. I don't like it.'' The RNN correctly handles the intensifier and negation context.}
    \label{fig:recap}
    \end{center}
    \vskip -0.2in
\end{figure*}

Our main contributions in this work are:\vspace{-0.15in}
\begin{itemize}[itemsep=-1.5mm]
    \item a data-driven method for identifying contextual inputs,
    \item a breakdown of the types of contextual effects learned,
    \item analysis of the strength and timescale of these effects,
    \item a mathematical model that describes this behavior,
    \item and a demonstration that simple and interpretable baseline models, augmented to incorporate this new understanding, recover almost all of the accuracy of the nonlinear RNNs.
\end{itemize}

%% file: text/02-preliminaries.tex
\section{Background}

Previous work~\cite{maheswaranathan2019b} analyzed recurrent networks trained to perform sentiment classification. They found that RNNs learned to store the current prediction as the location along a 1D manifold of approximate fixed points of the dynamics, called a \textit{line attractor}~\cite{Seung1996,Mante2013}. Additionally, they found that words with positive and negative valence drive the hidden state along this line attractor, which is aligned with the readout. A recap of this analysis is presented in Figure \ref{fig:recap}a, for a gated recurrent unit (GRU)~\cite{Cho2014} trained using the Yelp 2015 dataset~\cite{Zhang2015}. 

Critically, the mechanisms discussed in~\citet{maheswaranathan2019b} involved only integration of valence tokens along the line attractor. We tested the RNN on the same task after randomly shuffling the input words. Shuffling breaks apart important contextual phrases, such as ``really awesome'' and ``not bad''. On the shuffled test examples, the accuracy drops by $\sim\!4\%$ (Fig.~\ref{fig:supp-arch}a), close to the performance of a Bag-of-Words (BoW) baseline model. Thus, there are additional accuracy gains, present in the best performing RNNs, that are not explainable using prior known mechanisms.

The capability of RNNs to understand context can also be observed by probing the RNN with examples that contain contextual effects. Figure~\ref{fig:recap}b shows the predictions of the RNN in response to three probe reviews, a baseline (``This movie is awesome. I like it.''), one with \emphasis{intensifiers} (``This movie is \emphasis{extremely} awesome. I \emphasis{definitely} like it.''), and one with \negation{negation} (``This movie is \negation{not} awesome. I \negation{don't} like it.''). The RNN is capable of correctly assessing the sentiment in these reviews. Below, we break down exactly how the RNN is able to accomplish this.

\section{Preliminaries}

\subsection{Linearization and expansion points}
We denote the hidden state of a recurrent network at time $t$ as a vector, $\bh_t$. Similarly, the input to the network at time $t$ is given by a vector $\bx_t$. We use $F$ to denote a function that applies the recurrent network update, i.e. $\bh_{t} = F(\bh_{t-1}, \bx_t)$. The RNN defines an input-driven discrete-time dynamical system that sequentially processes inputs, in this case words in a document encoded as a sequence of one-hot input vectors. The final prediction (logit) is an affine projection (or readout) of the hidden state. In this work, we focus on binary sentiment classification, thus the logit is a scalar ($\bw^T  \bh_{t} + b$), with readout weights $\bw$ and bias $b$.

We can write the first-order approximation to the RNN dynamics~\cite{Khalil2001,maheswaranathan2019b} around an expansion point ($\bh^\te$, $\bx^\te$) as:
\begin{equation}
    \bh_{t} \approx F(\bh^\te, \bx^\te) + \Jrec\evalat{(\bh^\te, \bx^\te)} \Delta \bh_{t-1} + \Jinp\evalat{(\bh^\te, \bx^\te)} \Delta \bx_t, \label{eq:lin}
\end{equation}
where $\Delta \bh_{t-1} = \bh_{t-1}\!-\!\bh^\te$, $\Delta \bx_t = \bx_t\!-\!\bx^\te$, and $\{\Jrec, \Jinp\}$ are Jacobian matrices computed at the expansion point. In particular, the \textbf{recurrent Jacobian} $\left(J^\text{rec}_{ij}\evalat{(\bh^\te, \bx^\te)} = \frac{\partial F(\bh, \bx)_i}{\partial h_j}\right)$ defines the recurrent local dynamics and the \textbf{input Jacobian} $\left(J^\text{inp}_{ij}\evalat{(\bh^\te, \bx^\te)} = \frac{\partial F(\bh, \bx)_i}{\partial x_j}\right)$ defines the system's sensitivity to inputs.

\subsection{Linearization to understand dynamics}
\textbf{Fixed points} are points in state space that remain the same when applying the RNN: $\bh^* = F(\bh, \bx^*)$.
If the expansion point $(\bh^\te, \bx^\te)$ is an approximate fixed point of the dynamics, equation (\ref{eq:lin}) simplifies to a \textit{linear} dynamical system:
\begin{equation}
    \Delta \bh_{t} \approx \Jrec\evalat{(\bh^*, \bx^*)} \Delta \bh_{t-1} + \Jinp\evalat{(\bh^*, \bx^*)} \Delta \bx_t. \label{eq:lin_alds}
\end{equation}
We use equation (\ref{eq:lin_alds}) to study integration dynamics around approximate fixed points, as in ~\citet{maheswaranathan2019b}. These approximate fixed points are found numerically (see Supp. Mat. \S\ref{supp:fps} for details).
 
\subsection{Linearization to understand modifier words} 
To study the effects of modifier tokens, we also need to analyze the system \textit{away} from approximate fixed points.
In particular, we analyze the system at $\hmod$, defined as the state after processing a particular modifier word.
To do this, we linearize with respect to just the inputs for a single time step, expanding only in $\bx$, around $\bx^\te=\bzero$:
\begin{equation}
    \Delta \bh_{t} = \bh_t - F(\hmod, \bzero) \approx \Jinp\evalat{(\bh^{\text{mod}}, \bzero)} \bx_t, \label{eq:lin_inp}
\end{equation}
where equation (\ref{eq:lin_inp}) does not expand in $\bh$, to focus on input sensitivity. We make extensive use of the input Jacobian in equation (\ref{eq:lin_inp}) in our analysis in \S\ref{results:main}.

%% file: text/03-synthetic.tex
\section{Toy language to isolate modifier dynamics}
\begin{figure}[ht]
    \begin{center}
    \centerline{\includegraphics[width=0.7\columnwidth]{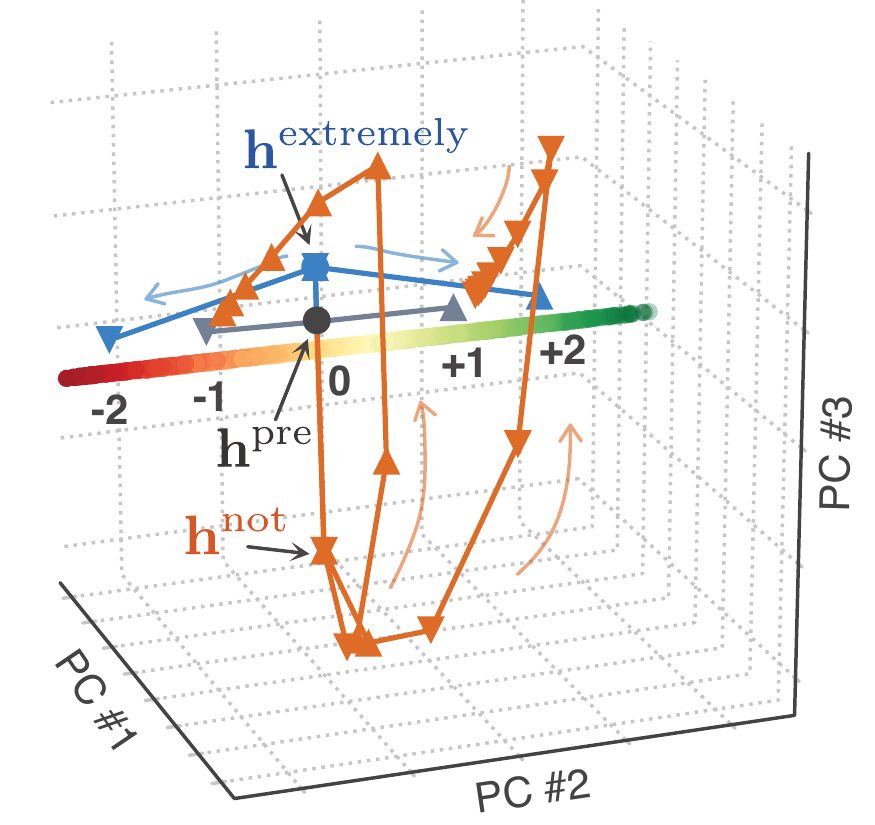}}
    \caption{State-space plot of a GRU trained on a toy language to isolate effects of modifier tokens. This example shows a line attractor (line from red to green) that performs integration of valence (value from -2 to +2, bold black numbers). Modifiers such as ``\emphasis{extremely}'' (blue) and ``\negation{not}'' (orange) achieve their effects by deflecting the state away from the line attractor to \emphasis{$\bh^{\text{extremely}}$} and \negation{$\bh^{\text{not}}$}, respectively. The effect of these deflections is to modify the valence of the input when projected onto the readout. Six trajectories are shown, all starting from $\bh^{\text{pre}}$. Gray shows the single step trajectory for ``good'' (\textcolor{gray}{$\blacktriangle$}, +1) and ``bad'' (\textcolor{gray}{$\blacktriangledown$}, -1). In blue are the single-step trajectories for ``\emphasis{extremely} good'' (\textcolor{blue}{$\blacktriangle$}, +2) and ``\emphasis{extremely} bad'' (\textcolor{blue}{$\blacktriangledown$}, -2), and the multi-step trajectories for ``\negation{not} good'' (\textcolor{orange}{$\blacktriangle$}, -1) and ``\negation{not} bad'' (\textcolor{orange}{$\blacktriangledown$}, +1) are in orange. We defined the duration of the effect of the ``\negation{not}'' modifier to be 4 tokens.}
    \label{fig-synth-intro}
    \end{center}
    \vskip -0.2in
\end{figure}

In order to illuminate how negation or emphasis is implemented in RNNs we developed a small toy language to isolate modifier effects. The language consisted of a small number of valence tokens, each with integer valence $\{-2, -1, 0, 1, 2\}$, analogous to words such as ``awful'', ``bad'', ``the'', ``good'', and ``awesome'', respectively. In addition, we added two modifier words, an intensifier that doubled the valence of the next input, and a negator that flipped the sign of the valence of the next four inputs (analogous to words such as ``\emphasis{extremely}'' and ``\negation{not}'').

We generated reviews using this language by  randomly ordering the tokens and trained RNNs to track the corresponding sentiment, defined as the cumulative sum of potentially modified valence inputs. For example, after training, the RNN correctly integrated ``good'' as +1, ``\emphasis{extremely} good'' as +2 and ``\negation{not} good'' as -1. We analyzed the networks using the methods developed by~\citet{maheswaranathan2019b}. An example state-space plot from a trained network is shown in Figure \ref{fig-synth-intro}. Note that this RNN also exhibits line attractor dynamics.

We draw two key insights from this exercise (see Supp. Mat. \S\ref{supp:toy} for a full analysis). First, modifiers achieve their effects by deflecting the state \textit{away} from the line attractor as opposed to along it, the latter of which is what valence words do. We found that the effect of this deflection away from the line attractor on the valence of the subsequent word was very well approximated by $\bw^T\;\Jinp\evalat{(\hmod,\textbf{0})}\;\bx^\text{val}$ (Supp. Fig. \ref{fig:toy-inp-jac-approx}), where $\hmod$ is the state after the modifier input, $\bx^\text{val}$ is the valence word following the modifier, and $\bw$ and bias $b$ are the readout weights and bias. As shown in the Supp. Mat. \S\ref{supp:toy}, removing the projection of $\Jinp\evalat{(\bh^{\mbox{\tiny{mod}}},\textbf{0})}\;\bx$ into the modification subspace removes the effect of the modifier (Supp. Fig. \ref{fig:toy-mode-removal}).

Second, the deflection of the state away from the line attractor caused by $\bx^{\text{mod}}$ remains for the duration of the modifier effect. For example, in the toy language we confined the temporal extent of the modification effects of ``extremely'' and ``not'' to one word and four words, respectively, and the corresponding deflections off the line attractor remain for one and four time steps, respectively. Finally, the transient dynamics associated with valence modification can be isolated in the local linear dynamics around fixed points on the line attractor (Supp. Fig. \ref{fig:toy-eig-plots} \& \ref{fig:toy-fixed-point-mode-angle}).

%% file: text/04-results.tex
\section{Reverse engineering contextual processing}
\label{results:main}

We turn our attention now to natural language, studying our best performing RNN, a GRU, trained to perform sentiment classification on the Yelp 2015 dataset~\cite{Zhang2015}.

\subsection{Identifying modifier words}
\label{results:indentifying_mods}
\begin{figure}[ht]
    \begin{center}
    \centerline{\includegraphics[width=0.7\columnwidth]{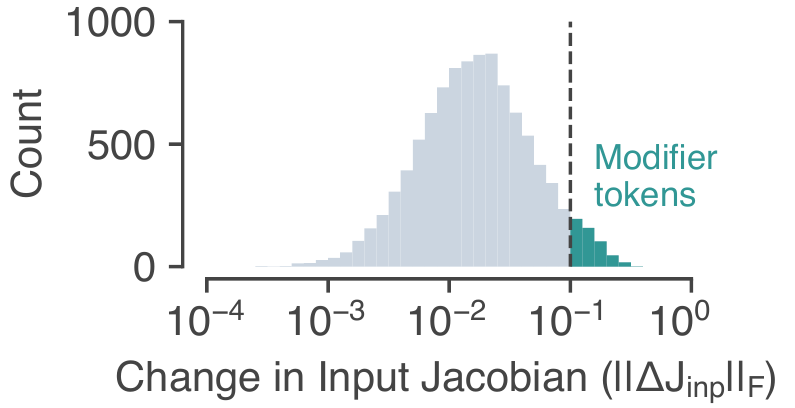}}
    \caption{Identifying modifier words. Histogram of Frobenius norm of change in input Jacobian. Note the log scaling on the x-axis, implying this distribution is heavy-tailed. We defined a modifier token as anything with a norm greater than 0.1. Example words above this threshold that are intuitively modifiers include ``not'', ``never'', ``overall'', and ``definitely''. Additional words found by this measure that are not as intuitive include: ``zero'', ``two'', ``poisoning'', and ``worst''. }
    \label{fig:dij}
    \end{center}
    \vskip -0.2in
\end{figure}

For RNNs trained on real world datasets, we did not know what words might act as modifiers, therefore, we wanted a data driven approach to identify them. Inspired by the toy model, we looked for particular words that deflected the hidden state to dimensions where the input Jacobian changed substantially, as that indicated differential processing of inputs. We defined the change in input Jacobian after a particular input, $\bx^{\text{mod}}$, as:
\begin{equation}
  \Delta \Jinp\rvert_{\hmod} \equiv \frac{dF}{d\bx}\biggr\rvert_{(\hmod, \bzero)} - \frac{dF}{d\bx}\biggr\rvert_{(\bh^{*}, \bzero)}, \label{eq:DJinp}
\end{equation}
where $\hmod = F(\bh^*, \bx^{\text{mod}})$ is the hidden state after processing $\bx^{\text{mod}}$, starting from a point on the line attractor, $\bh^*$\footnote{We suppress the $\bh^*$ in the notation as it was held constant in our analyses.}. Thus, $\Delta \Jinp\rvert_{\hmod}$ measures how the system's processing of words changes as a function of a a preceding word, $\bx^{\text{mod}}$.

We studied the size of the changes in input processing by computing the Frobenius norm, $\|\Delta \Jinp\rvert_{\hmod}\|_F$, for each of the top 10,000 most common words in the dataset. The resulting distribution is shown in Figure \ref{fig:dij} and is evidently heavy-tailed. Examining words with large values reveals common negators (e.g. ``not'', ``never'') and intensifiers (e.g. ``very'', ``definitely''). We selected a set of modifier words for further analysis by keeping words whose change in input Jacobian was greater than an arbitrary threshold of 0.1\footnote{None of the presented results are particularly sensitive to this choice.}.

\begin{figure}[ht]
    \begin{center}
    \centerline{\includegraphics[width=0.75\columnwidth]{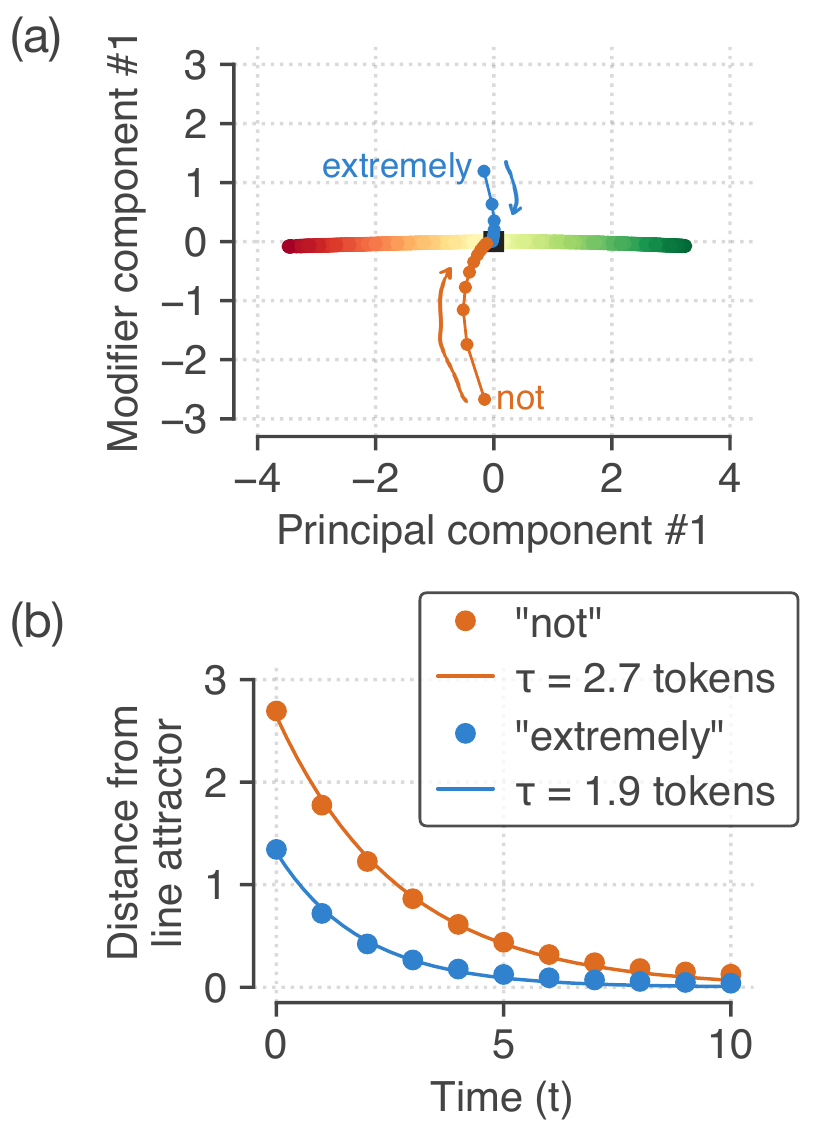}}
    \caption{Impulse response to a modifier word. (a) Two trajectories of the network hidden state in response to modifier words ``\emphasis{extremely}'' and ``\negation{not}'', projected onto the top PCA component (x-axis) as well as the top modifier component (y-axis); see text for details. (b) Distance from the line attractor after a single modifier word. Circles show response of the RNN, line is an exponential fit.}
    \label{fig:impulse}
    \end{center}
    \vskip -0.2in
\end{figure}

\subsection{Analyzing example modifier words}
\label{results:example_mods}
\subsubsection{Dynamics of example modifier transients}

We next looked at the dynamics of the hidden state in response to example modifier words, ``\emphasis{extremely}'' or ``\negation{not}'', followed by a series of pad tokens. Figure \ref{fig:impulse}a shows the impulse response to ``\emphasis{extremely}'' and ``\negation{not}'', projected into a two-dimensional subspace for visualization\footnote{The two-dimensional subspace in Fig \ref{fig:impulse} consists of the top PCA component (x-axis) and the top modifier component (defined later in \S\ref{sec:modsubspace}).}. Each modifier word deflects the hidden state off of the line attractor, which then relaxes back over the course of 5-10 time steps, similarly to the toy example. Figure \ref{fig:impulse}b shows the same impulse response as a function of time, quantified using the Euclidean distance from the line attractor. These transients, induced by modifiers, explore parts of the RNN state space that critically are not contained in the top two PCA dimensions (Fig \ref{fig:recap}a). It was only through focusing on the change in input Jacobian that we were able to identify them.

\begin{figure}[ht]
    \begin{center}
    \centerline{\includegraphics[width=0.7\columnwidth]{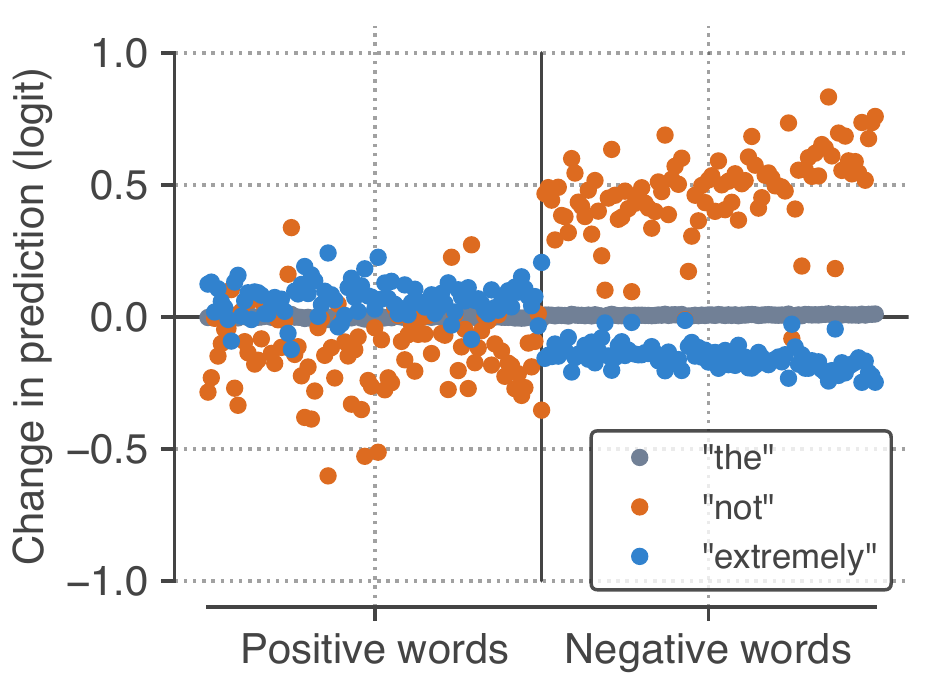}}
    \caption{Barcodes for visualizing the effects of modifiers. The barcode is a quantitative signature of a particular modifier word. Shown are the barcodes for `the' (gray), ``\emphasis{extremely}'' (blue), and ``\negation{not}'' (orange). Each point in a barcode is the change in the model's prediction in response to a particular valence word when a modifier word precedes it (e.g. ``not great'', ``not bad'', ``not amazing'', etc. (orange)), for the top 100 positive words (e.g. ``amazing'', ``awesome''; left) and the top 100 most negative words (e.g. ``hate'', ``awful''; right). For example the ``not'' barcode shows a negative or reduced change for positive valence words and a strong positive output change for negative words.}
    \label{fig:barcodes}
    \end{center}
    \vskip -0.2in
\end{figure}

\subsubsection{Modifier barcodes}
Next, we asked \textit{how} the system leverages these transient dynamics to enable contextual processing. To answer this, we analyzed how the input Jacobian changes along these transients in comparison to their values at the line attractor. However, a complication arises because a modifier may have differing effects on each word and there are many words in the vocabulary. Therefore we developed a visualization, a \textbf{modifier barcode}, to study the effect of changing Jacobians.  Barcodes are intended to quantify how a given modifier word affects processing of future inputs. For a given probe word $\bx$, we define a barcode for a modifier word as:
\begin{align}
    \text{barcode}^{\text{mod}}(\bx) &= \bw^T \Delta \Jinp\evalat{\bh^{\text{mod}}}\; \bx, \label{eq:barcode}
\end{align}
where $ \bh^{\text{mod}}$ is the RNN state after a modifier word has been processed. As there were 90,000 words in the vocabulary we instead selected 100 positive and 100 negative words (e.g. ``awesome'' and ``awful'') to comprise the barcode\footnote{Words were selected by taking the 100 largest (most positive) and 100 smallest (most negative) logistic regression weights, using a separately fit logistic regression classifier (Bag-of-words model). Results do not depend on the particular set of probe words chosen.}.

Figure \ref{fig:barcodes} shows three barcodes, for the words ``the'', ``\emphasis{extremely}'', and ``\negation{not}''. For words that are not modifiers (such as ``the''), the effects of inputs before and after the word are the same, thus the barcode values are close to zero. For intensifiers, such as ``\emphasis{extremely}'', we see that positive words (left half of Figure \ref{fig:barcodes}) and negative words (right half) are accentuated). For negators, such as ``\negation{not}'', positive and negative words are made relatively more negative or more positive, respectively. In summary, the modifier barcode allows us to easily assess \textit{what} the effects of a particular modifier are, through the changing input Jacobian.

\subsection{Summary across all modifier words}
\label{results:summary}

\begin{figure}[ht]
    \begin{center}
    \centerline{\includegraphics[width=0.75\columnwidth]{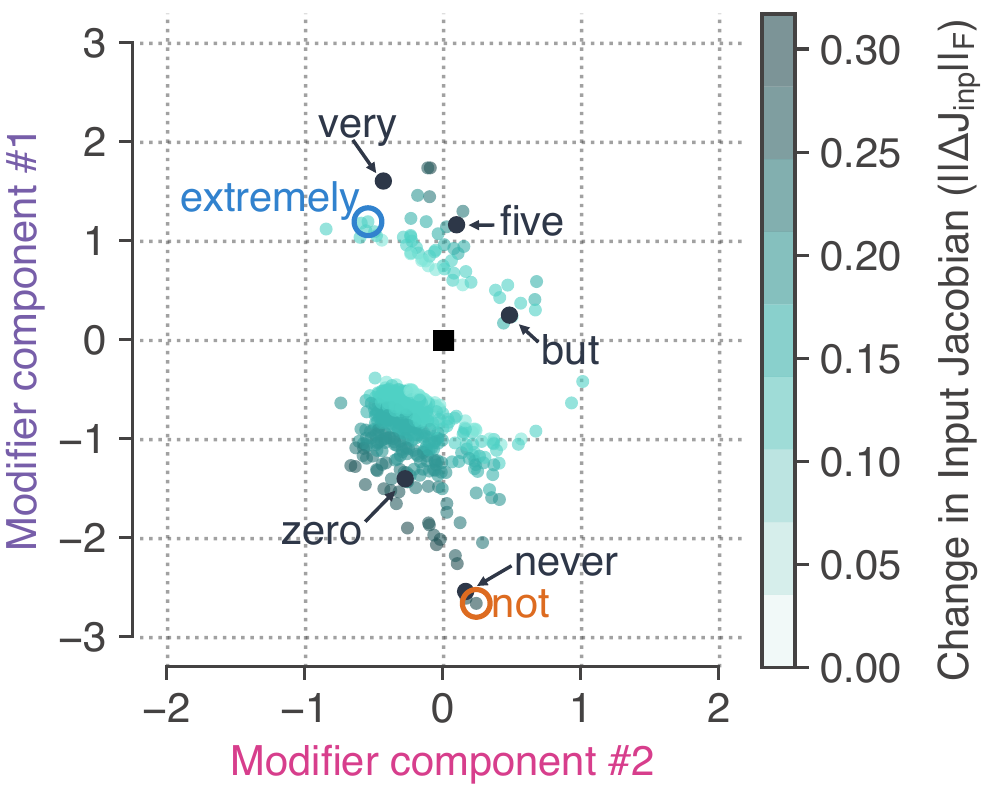}}
    \caption{Low-dimensional modifier subspace. Projection of modifier points (teal circles) onto the top two modifier components, obtained by performing principal components analysis (PCA) on the set of states after processing modifier words. Highlighted are a number of example modifiers, including ``\negation{not}'' and ``\emphasis{extremely}'') from the previous figure. Emphasizers (e.g. ``extremely'', ``very'') are on one side of the subspace, whereas negators (e.g. ``not'', ``never) are on the other side.}
    \label{fig:subspace}
    \end{center}
    \vskip -0.2in
\end{figure}

\subsubsection{Modifier subspace}
\label{sec:modsubspace}
So far we have analyzed the modifier dynamics and barcodes for a couple of example modifiers. Next, we will summarize these effects over the entire set of identified modifier words. In particular, we are interested in understanding how many different types of modifiers are there, and whether we can succinctly describe their effects and dynamics.

First, we collected all of the deflections in the hidden state induced by modifier words. This is a set of points in the RNN hidden state where there are substantial changes to the input Jacobian. We ran principal components analysis on this set of points to identify a low-dimensional \textit{modifier subspace}, also ensuring that the subspace is orthogonal to the line attractor\footnote{We do this additional orthogonalization step to ensure that the identified modifier subspace does not have any valence effects, which occur along the line attractor.}. In particular, two components explained 96.2\% of the variance, suggesting that the dominant modifier effects can be understood as occurring in a 2D plane.

We projected all of the modifier inputs into this 2D subspace for visualization to highlight the distribution of modifiers in this subspace, as well as some example modifiers (Fig. \ref{fig:subspace}). Negators (words like ``not'' and ``never'') live in one part of the subspace, whereas intensifiers (words like ``extremely'' and ``definitely'') live in another part. Interestingly, these two types of modifiers share the same subspace. %

\subsubsection{Fast and slow modifier dynamics}

\begin{figure}[ht]
    \begin{center}
    \centerline{\includegraphics[width=0.7\columnwidth]{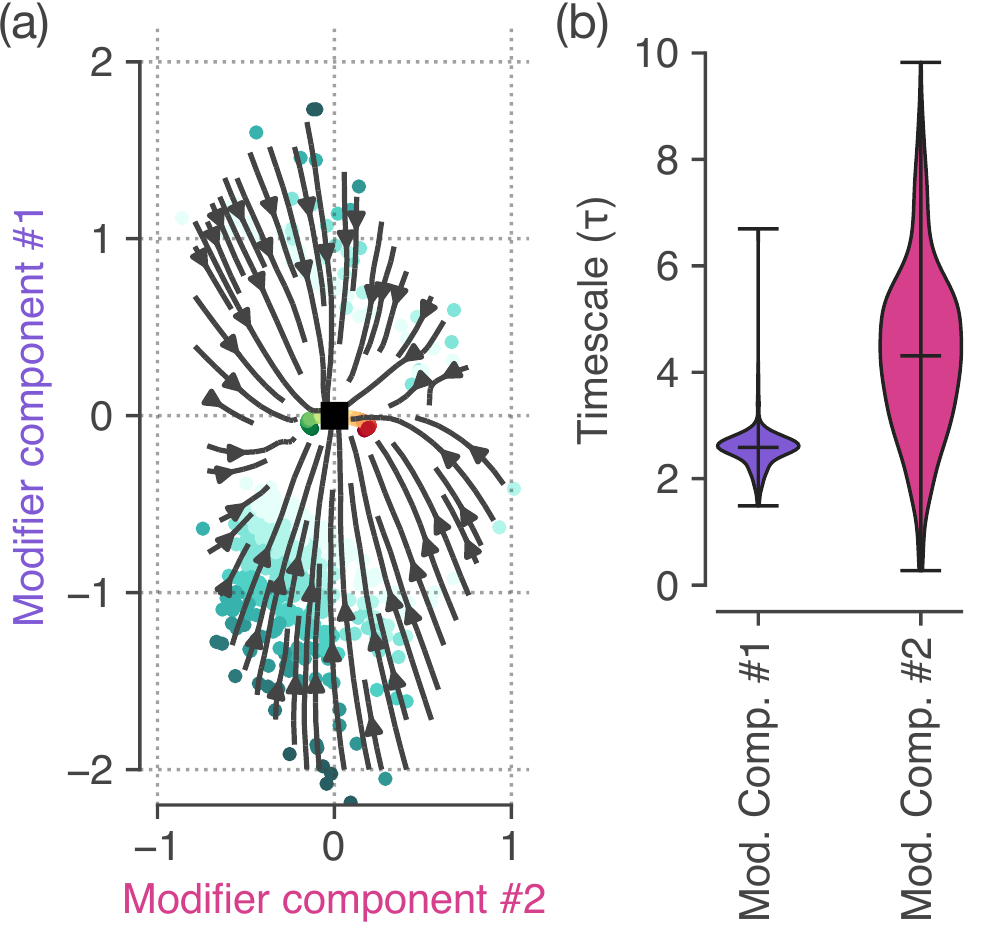}}
    \caption{Timescales of modifier effects. (a) Hidden state dynamics of the impulse response to modifier inputs. Note the line attractor projects out of the page. (b) Distribution of the estimated timescale of the decay along each modifier component across modifier words.}
    \label{fig:dynamics}
    \end{center}
    \vskip -0.2in
\end{figure}

We then studied the dynamics in the modifier subspace. In particular, the dynamics of how the hidden state evolves in this subspace determines the length of the effects of each modifier (called the ``scope'' in linguistics \cite{quirk1985,horn2000negation}). Figure \ref{fig:dynamics} shows the dynamics of modifiers in the 2D subspace, along with a quantification of the timescale of the decay in each modifier dimension or component (Fig. \ref{fig:dynamics}b). We observed that the first modifier component is faster (mean $\tau$ = 2.6 tokens), whereas the second one is slower (mean $\tau$ = 4.3 tokens), though there is spread in the distributions in both dimensions. This argues that the scope of modifier words in RNNs trained to perform sentiment classification lasts for tens of tokens and agrees with estimates of the scope of negation from human annotators \cite{taboada2011lexicon,strohm2018empirical}.

\begin{figure}[ht]
    \begin{center}
    \centerline{\includegraphics[width=0.7\columnwidth]{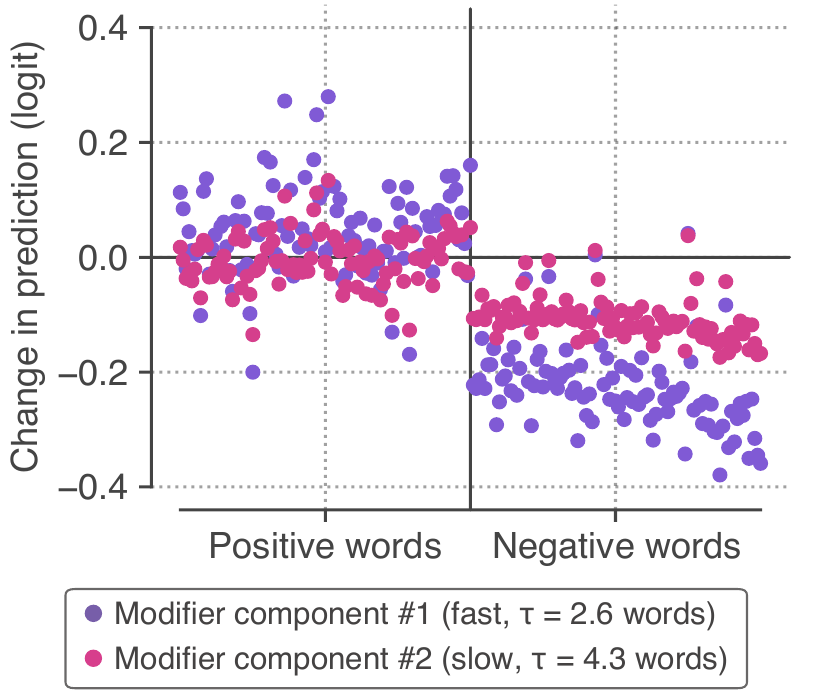}}
    \caption{Barcodes for the principal components (eigenvectors) of the modification subspace. These barcodes show how the two dimensions of the modification subspace support rich modification of valence words with two time constants.}
    \label{fig:bilinear_barcodes}
    \end{center}
    \vskip -0.2in
\end{figure}

\subsubsection{Summary of modifier barcodes}
Beyond dynamics, we were also interested in how the input Jacobian changes in this two dimensional subspace. In particular, we wanted to understand whether or not there are different types of modification effects. To look at this, we computed barcodes for the top two modifier axes\footnote{Although we focus on the top two components in the main text, the barcodes for the top 10 modifier components are presented in Supp. Fig.~\ref{fig:additional-barcodes}.} (Fig. \ref{fig:bilinear_barcodes}). We saw that qualitatively, the two dimensions have similar effects, so the predominant difference between them is presumably the difference in timescales discussed above.

However, there are still rich patterns of effects in terms of the barcode weights for individual valence words. For example, the barcode weight corresponding to the valence word ``stars'' has a strong projection on the first modifier component, but not the second. This suggests that modifier words that induce important contextual changes for ``stars'' (e.g. ``zero stars'' and ``five stars'') should preferentially have larger projections onto the first modifier component, and not the second. Indeed, we see that ``zero'' and ``five'' are included in the set of modifier words, and have projections only on the first, faster modifier component~(Fig. \ref{fig:subspace}). By tuning the input Jacobian for different words along these two modifier components, the network is able to correspondingly tune the timescale of their effects.

\subsubsection{Perturbation experiment}
Finally, to test whether the modifier subspace is necessary for contextual computation, we performed a perturbation experiment.
To do this, we evaluated the network as usual except at every timestep we projected the hidden state out of the modifier subspace. Doing this, we reasoned, should interrupt contextual processing, but leave integration of valence intact. Indeed, we find that this perturbation has these expected effects both on single examples and across all test examples (Supp. Fig.~\ref{fig:perturbation}, \S\ref{supp:perturbations}).

\section{Other contextual effects}
In addition to modifiers, we found contextual effects that are active during network processing of the beginning and end of documents.

\begin{figure*}[ht]
    \begin{center}
    \centerline{\includegraphics[width=0.95\textwidth]{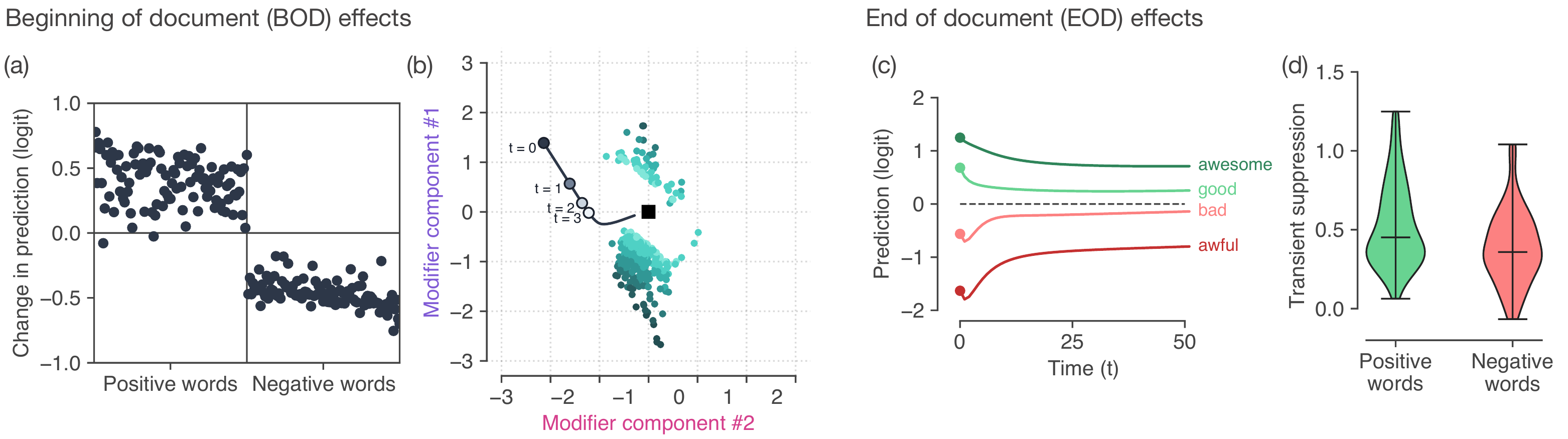}}
    \caption{Contextual effects that occur at the beginning (a-b) and end (c-d) of documents. (a) Barcode measured at the initial state, $\bh_{\bzero}$, reveals that the initial state emphasizes sentiment. (b) Projection of the (trained) initial state (gray circle, $t=0$) onto the modifier subspace and corresponding dynamics in response to 50 pad tokens (gray line, $t=1,\ldots,50$). The initial state initially has a large projection onto both modifier components, which means that words in the beginning of each document are \textit{emphasized} (weighted more strongly) relative to words in the middle of the document. (c) When a valence word enters the network it introduces a transient deflection on the readout that decays in $\sim\!50$ steps. This means that words at the end of a review are modified, as the transient has not had time to decay. In the four examples shown, the effect is to emphasize these words.  (d) Summary of the amount of transient suppression across many words.}
    \label{fig:boseos}
    \end{center}
    \vskip -0.2in
\end{figure*}

\subsection{Beginning of documents}
After training, we found that the RNN's initial state, $\bh_{\bzero}$ (a vector of trainable parameters) is far from the line attractor (Fig. \ref{fig:recap}a), which surprised us. The reason for this was revealed when we computed the modifier barcode corresponding to the learned initial state (Fig. \ref{fig:boseos}a). This barcode has a signature very much like that of an intensifier word such as ``extremely'', in that it accentuates both positive and negative valence. Moreover, we can also see that $\bh_{\bzero}$, when projected into the modifier subspace, has a significant projection on both modifier components (Fig. \ref{fig:boseos}b). The implication of this is that words at the beginning of a document will be emphasized relative to words later in the document. 

Presumably, the RNN learned to do this because it improves training performance. We tested this hypothesis directly using a perturbation experiment. We projected $\bh_{\bzero}$ out of the 2D modifier subspace, and then re-tested the GRU using this new initial state. We found that this perturbation caused a drop in accuracy of 0.17\% on the test set and 0.23\% on the train set, corresponding to an decrease in 68 and 1150 correct reviews, respectively. Importantly, projecting out a \textit{random} 2D subspace did not affect the accuracy, indicating that this accuracy effect is unique to the modifier subspace.

\subsection{End of documents}
We also found contextual effects that emphasized words at the end of documents. Here, we identified a different mechanism responsible: short-term transient dynamics. When we looked at the impulse response of the system (Fig. \ref{fig:boseos}c), we found that the projection onto the readout in response these tokens is initially large, but then decays over the course of around 50 tokens, until it settles to the steady-state valence of each word. We quantified this effect by computing the ratio of the steady-state valence to its initial valence. We did this for a large set of positive and negative words~(Fig. \ref{fig:boseos}d), and find consistent effects. The implication is that words at the end of a review are also emphasized, as their transient effects do not have time to decay away. We further identified two linear modes of the recurrent dynamics responsible for these transient effects (analyzed in Supp. Mat. \S\ref{supp:eod}).

We also tested that end of document emphasis matters to performance with another perturbation experiment. We added 50 pad tokens to the end of each review, thus allowing the transient dynamics to decay and thereby removing their effects along the readout. We evaluated the RNN on this padded dataset and found that the test accuracy dropped by 0.05\% and the train accuracy by 0.2\%, corresponding to incorrectly categorizing 20 and 1000 reviews, respectively.

\section{Additional RNN architectures}
\begin{figure*}[ht]
    \begin{center}
    \centerline{\includegraphics[width=0.9\textwidth]{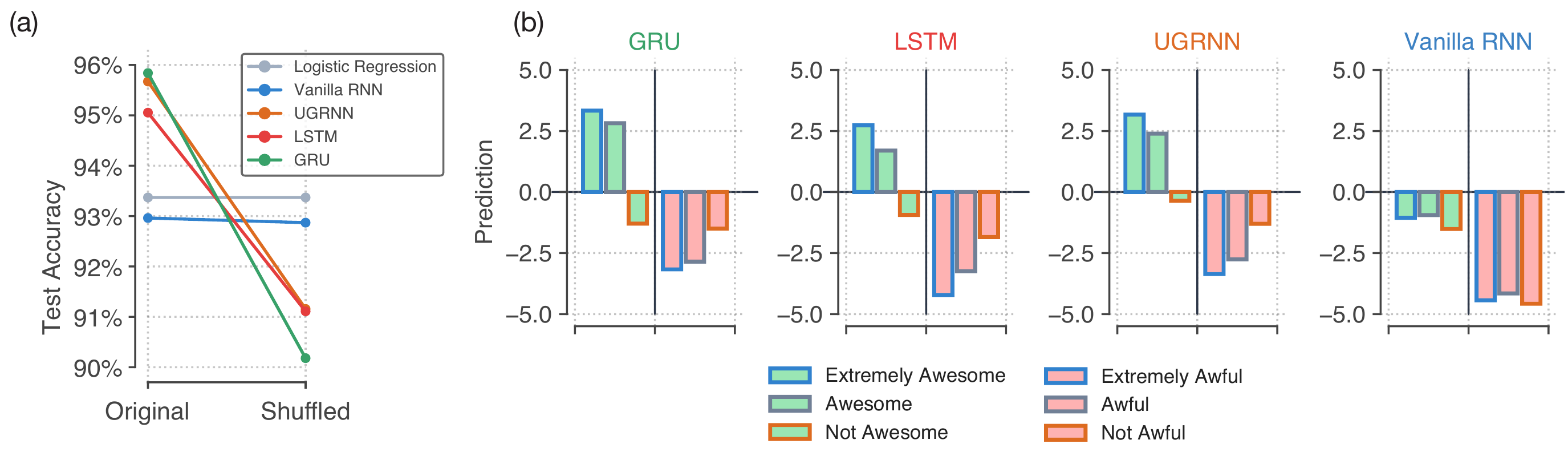}}
    \caption{Summary across different architectures. (a) We performed a control experiment by randomly shuffling the tokens of each example sequence. We then classified these shuffled reviews using the RNNs trained on the original (non-shuffled) data. These results show that word order matters for gated RNNs but not for the Vanilla RNN or BoW model; thus the gated RNNs take word order into account. (b) Example predictions for RNNs for the phrases ``\emphasis{extremely} awesome'', ``awesome'', ``\negation{not} awesome'', ``\emphasis{extremely} awful'', ``awful'', and ``\negation{not} awful''. The gated RNN architectures show similar, modulated responses to these tokens while the Vanilla RNN shows little modulation of the valences of either ``awesome'' or ``awful'', regardless of the preceding modifier word.}
    \label{fig:supp-arch}
    \end{center}
    \vskip -0.2in
\end{figure*}

So far, we have analyzed a particular RNN architecture, a GRU. We additionally trained and analyzed other RNN architectures, including an LSTM~\cite{Hochreiter1997}, Update Gate RNN~\cite{Collins2016}, and a Vanilla RNN. We found that all of the gated architectures were capable of processing modifiers. In particular, shuffling inputs reduced accuracy for the gated architectures by a similar amount  (Figure \ref{fig:supp-arch}a), and they have similar responses to test phrases with modifier words  (Figure \ref{fig:supp-arch}b). Finally, we repeated the analyses in Figures 3-6 for the LSTM, which are shown in Supp. Mat. \S\ref{supp:lstm}. This hints at universal mechanisms underlying contextual processing in gated RNNs~\cite{maheswaranathan2019a}.

\section{Bilinear model captures modifier effects}\label{bilinear}

Taken together, our analyses reveals the mechanisms by which RNNs perform contextual processing.
In particular, the key mechanism involves changing the input Jacobian ($\Jinp$) in the modifier subspace, and modifier words explicitly project the state into that subspace.

We can synthesize \textit{how} the input Jacobian changes by expressing it as the sum of the input Jacobian at the line attractor, $\Jinp\evalat{(\bh^*, \bzero)}$, with $P$ additional bilinear terms:
\begin{equation}
  \Jinp\evalat{(\bh^{\text{mod}}, \bzero)} = \Jinp\evalat{(\bh^*, \bzero)} + \sum_{p=1}^P \left(( \bh^\text{mod}\!-\!\bh^*)^T \mathbf{m}^{p}\right) \mathbf{A}^p, \label{eq:bilinear}
\end{equation}
where $\mathbf{A}^p$ is a fixed matrix (with the same dimensions as the input Jacobian) and $\bm^p$ is a modifier component.
Thus we capture the variation in $\Jinp$ as a function of $\bh^{\text{mod}}$ by starting with the input Jacobian at $\bh^*$, a point along the line attractor, and adding a small number of bilinear terms to it.

The additional $P$ terms consist of two components. The first, $( \bh^\text{mod}\!-\!\bh^*)^T \mathbf{m}^{p}$, a scalar, captures the strength or amount of modification as a function of the projection of the hidden state onto the modifier component $\mathbf{m}^p$. As the hidden state decays back to the line attractor, the modification effects naturally decay along with the dot product $(\bh_t\!-\!\bh^*)^T \mathbf{m}^p$, with $\bh_0 = \bh^\text{mod}$. The second term, $\mathbf{A}^p$, models the corresponding change in the input Jacobian. These $\mathbf{A}^p$ terms capture variation in the barcodes in Figure \ref{fig:bilinear_barcodes}. We fit the parameters ($\bm^p$ and $\mathbf{A}^p$) by regressing eq. (\ref{eq:bilinear}) against the actual input Jacobian for the set of modifier words\footnote{This regression can be solved in closed form via a singular value decomposition (SVD).}. This approximation captures 83\%, 96\%, and 98\% of the input Jacobian variation across all modification tokens for $P$ $\in [1,2,3]$, respectively.

\section{Augmented baseline models with insights from RNNs recover performance} \label{baselines}
The result that the bilinear model defined in equation (\ref{eq:bilinear}) achieves 98\% of the variance in $\Jinp\evalat{(\bh^{\text{mod}}, \bzero)}$ with only $P=3$ terms motivates baseline models of the form 
\begin{align}
    \bw^T \Jinp\evalat{(\bh^*, \bzero)} +  \sum_{p=1}^P  \left((\bh^\text{mod}\!-\!\bh^*)^T \mathbf{m}^{p}\right) \bw^T \mathbf{A}^p + b, \label{eq:baseline_motivation}
\end{align}
where $\bw$ and $b$ are the readout weights and bias, respectively. First, we note that the first term $\bw^T \Jinp\evalat{(\bh^*, \bzero)} \in \mathbb{R}^W$ is analogous to the weights $\boldsymbol{\beta}\in \mathbb{R}^W$ in a Bag-of-Words model (BoW). The additional $P$ terms can be interpreted as additional $\boldsymbol{\beta}_{\text{mod}}$ weights that activate after a modifier word occurs and whose effect decays as $\bh^\text{mod}$ returns to the line attractor. Thus, we propose augmented models of the form
\begin{align}
    \sum_{t=1}^T \left(\beta[t] + \sum^P_{p=1} \left(f^m \ast \mu^m\right)[t]\; \beta^p_{\text{mod}}[t]\right) + b, \label{eq:baseline}
\end{align} 
where we have transitioned to indexing $\boldsymbol{\beta}$ by time $t$, which selects the weight for the word occurring at index $t$ in the review. Modifier weights $\beta^p_{\text{mod}}[t]$ are defined separately from the baseline valence weights $\beta[t]$. 

We model the decaying effect of modifiers via convolution of an indicator that signals the location of each modifier word, $\mu^m[t]$, with a causal exponential filter $f^m[s] = \alpha^m \exp(-s/\tau^m)$, with scaling $\alpha^m$ and timescale $\tau^m$.

All of the augmented baseline models have the form given by equation (\ref{eq:baseline}), with different choices for what modifiers to include. In particular, we trained the following models (full descriptions are in Supp. Mat. \S\ref{supp:baselines}):
\begin{description}[itemsep=-1.5mm]
\item[Bag-of-Words (BoW):] Baseline model with no modifiers.
\item[Convolution of Modifier Words (CoMW):] BoW plus convolutional modifiers with the \textit{same} regression weights (modifiers do not have unique weights).
\item[Convolution of BOD/EOD:] BoW plus two additional modifier tokens for the beginning and end of the document (BOD/EOD), with the same regression weights.
\item[CoMW + $\beta_\text{mod}$:] BoW plus conv. modifiers with \textit{learned} regression weights (modifiers have unique weights).
\item[Convolution of BOD/EOD + $\beta_\text{BOD} + \beta_\text{EOD}$:] BoW plus BOD/EOD modifiers, with learned regression weights.
\item[CoMW + $\beta_\text{mod} + \beta_\text{BOD} + \beta_\text{EOD}$:] BoW plus convolutional \textit{and} BOD/EOD modifiers, with learned regression weights (the most powerful of the above models).
\end{description}

For all of our baseline experiments we set the number of modifiers, $M=400$, and the number of modifier $\boldsymbol{\beta}_{\text{mod}}$ vectors to $P=3$ (as $P=3$ explains 98\% of the variance of $\Jinp\evalat{(\bh^{\text{mod}}, \bzero)}$ in the bilinear model in \S\ref{bilinear}).
The models were trained using the Adam optimizer~\cite{Kingma2014}. We selected hyperparameters (learning rate, learning rate decay, momentum, an $\ell_2$ regularization penalty, and dropout rate) via a validation set using random search. We found dropout directly on the input words to be a useful regularizer for the more powerful models.

Results for these augmented baselines are in Table \ref{tab:acc}.  The classic BoW model achieved 93.57\% (on the Yelp dataset) and represents a baseline that does \textit{not} implement contextual processing. The additional five baseline models range in increasing modeling power, with the most powerful baseline model achieving 95.63\%, a test accuracy that is very close to the best performing RNN (90\% of the difference).%

\begin{table}[ht]
    \centering
    \caption{Test accuracies across baseline models and RNNs for the Yelp and IMDB datsets.}
    \label{tab:acc}
    \begin{center}
    \begin{scriptsize}
    \begin{sc}
    \begin{tabular}{lrr}
         & Yelp 2015 & IMDB \\ \hline
         Baseline models &  &\\
         \hline
         \hline
         Bag of words & 93.57\% & 89.47\% \\
         Convolution of BOD \& EOD tokens & 93.99\% & 89.57\% \\
         Conv. + BOD \& EOD tokens + $\bbeta_{\mbor} + \bbeta_{\meor}$ & 94.37\% & 89.60\% \\ %
         Convolution of Modifier Words (CoMW) & 94.76\% & 89.90\% \\ %
         CoMW + $\bbeta_\text{mod}$ & 95.40\% & 90.68\% \\  %
         CoMW + $\bbeta_\text{mod} + \bbeta_{\mbor} + \bbeta_{\meor}$ & 95.63\% & 90.75\% \\ %
         \hline \\
         RNN models & \\
         \hline
         \hline
         GRU~\cite{Cho2014} & 95.84\% & 89.63\% \\
         LSTM~\cite{Hochreiter1997}& 95.05\% & 91.59\% \\
         Update Gate RNN~\cite{Collins2016}& 95.67\% & 89.43\% \\
         Vanilla RNN & 92.96\% & 89.55\% \\
        \bottomrule
    \end{tabular}
\end{sc}
\end{scriptsize}
\end{center}
\vskip -0.1in
\end{table}

%% file: text/05-related-work.tex
\section{Related Work}
Understanding context has long been an important challenge in natural language processing~\cite{morante2012modality,mohammad2017challenges}. In particular, including pairs of words (bigrams) as features in simple BoW models significantly improves classification performance~\cite{wang2012baselines}. \citet{polanyi2006contextual} introduced the idea of ``contextual valence shifters'', which model contextual effects by shifting valence for a hand crafted set of modifier words. Further work refined these ideas, demonstrating their usefulness in improving sentiment classification accuracy~\cite{kennedy2006sentiment,taboada2011lexicon}; identifying the amount and scope of the shifts combining human annotators~\cite{ruppenhofer2015ordering,schulder2017towards,kiritchenko2017effect} and automated methods~\cite{choi2008learning,ikeda2008learning,liu2009review,li2010sentiment,boubel2013automatic}; and finally using these lexicons to regularize RNNs~\cite{teng2016context,qian2016linguistically}. More recently, due to larger and more readily available datasets, neural networks achieve state-of-the-art performance on sentiment classification~\cite{dieng2016topicrnn,johnson2016supervised,Zhang2018}, \textit{without} explicitly building in contextual effects, often fine tuning larger models trained using unsupervised methods~\cite{howard2018universal,sun2019fine,yang2019xlnet}.

Methods for interpreting and understanding the computations performed by recurrent networks include: inspecting individual units~\cite{karpathy2015visualizing}; using input and structural probes~\cite{socher2013recursive,hewitt2019structural}; visualizing salient inputs~\cite{li2015visualizing,lime,arras2017explaining,Murdoch2018}, analyzing and clustering RNN dynamics~\cite{elman1990finding,elman1991distributed,Strobelt2016,ming2017understanding,maheswaranathan2019b}; and studying changes due to perturbed or dropped inputs~\cite{kadar2017representation}. For a review of these methods, and others, see \citet{belinkov2019analysis}. %

%% file: text/06-discussion.tex
\section{Discussion}

In summary, we provide tools and analyses that elucidate how neural networks implement contextual processing.

Here, we focused on unidirectional recurrent networks. Extending these tools to other tasks and architectures, such as those that use attention~\cite{bahdanau2014neural,vaswani2017attention}, is a promising future research direction.
For example, we predict that the reverse direction of a bidirectional RNN~\cite{schulder2017towards} would reveal backwards modifiers (e.g. the suffix \textit{-less} which has a suppressive effect on the preceding word stem, as in \textit{soulless}).

Our analysis reveals rich modifier effects and timescales learned by the RNN. These properties are remarkably consistent with properties of modifiers from the linguistics literature, including: the length or scope of contextual effects lasting a few words~\cite{chapman2001negation,taboada2011lexicon,reitan2015negation,strohm2018empirical}, the asymmetry in the strength and number of negators vs intensifiers~\cite{horn1989natural,kennedy2006sentiment,taboada2011lexicon,schulder2017towards}, the relative weighting of different intensifiers~\cite{ruppenhofer2015ordering}, and the fact that negation is better modeled as an additive effect rather than a multiplicative effect~\cite{zhu2014empirical}.

This speaks to the general scientific program of analyzing optimized universal dynamical systems to provide insights into the underlying structure of natural data.

%% file: text/07-acknowledgements.tex
\section*{Acknowledgements}

The authors wish to thank Alex H. Williams, Larry Abbott, Jascha Sohl-Dickstein, and Surya Ganguli for helpful discussions.

%% file: text/08-supplemental.tex
\section{Toy sentiment task}\label{supp:toy}
\subsection{Task introduction}
In order to illuminate how negation or emphasis is implemented in RNNs we developed a small toy language to isolate their effects. The language consisted of a small number of valence tokens, each with integer valence $\{-2, -1, 0, 1, 2\}$ (analogous to words such as ``awful'', ``bad'', ``the'', ``good'', ``awesome''). In addition, we added two modifier words, an intensifier that doubled the valence of the next input (analgous to words such as ``\emphasis{extremely}''), and a negator that reversed the valence of the next {\it four} inputs (analogous to words such as ``\negation{not}''). We varied the timescales of the modifier effects because there was evidence for this in the Yelp 2015 sentiment data we analyzed and also because we were interested in understanding modifier words with different dynamics.

We generated random reviews using this language by randomly selecting 50 words and placing them sequentially. To avoid overly complex effects we made sure that ``not'' could not follow ``not'' within 4 tokens and that ``extremely'' could not directly follow ``extremely''. We trained RNNs to track the corresponding sentiment, defined as the per-time step cumulative sum of potentially modified valence inputs. In this way, we could explicitly define per-time step target values for the output, thus controlling how an intensifier or negator works.  As we specified a per-time step integration target for the network, we used a least-mean-squares loss, as opposed to cross-entropy loss typically used for classification. These are differences from the standard sentiment classification setup where a review is only associated with a single positive or negative classification pertaining to the entire review. For example, after training, the RNN correctly integrated long sequences of words and we verified that the RNN integrated ``good'' as +1, ``\emphasis{extremely} good'' as +2 and ``\negation{not} the the the good'' as -1. 

\subsection{Confirmation of line attractor dynamics in the toy model}
\begin{figure}[ht]
\vskip 0.2in
\begin{center}
\centerline{\includegraphics[width=4in]{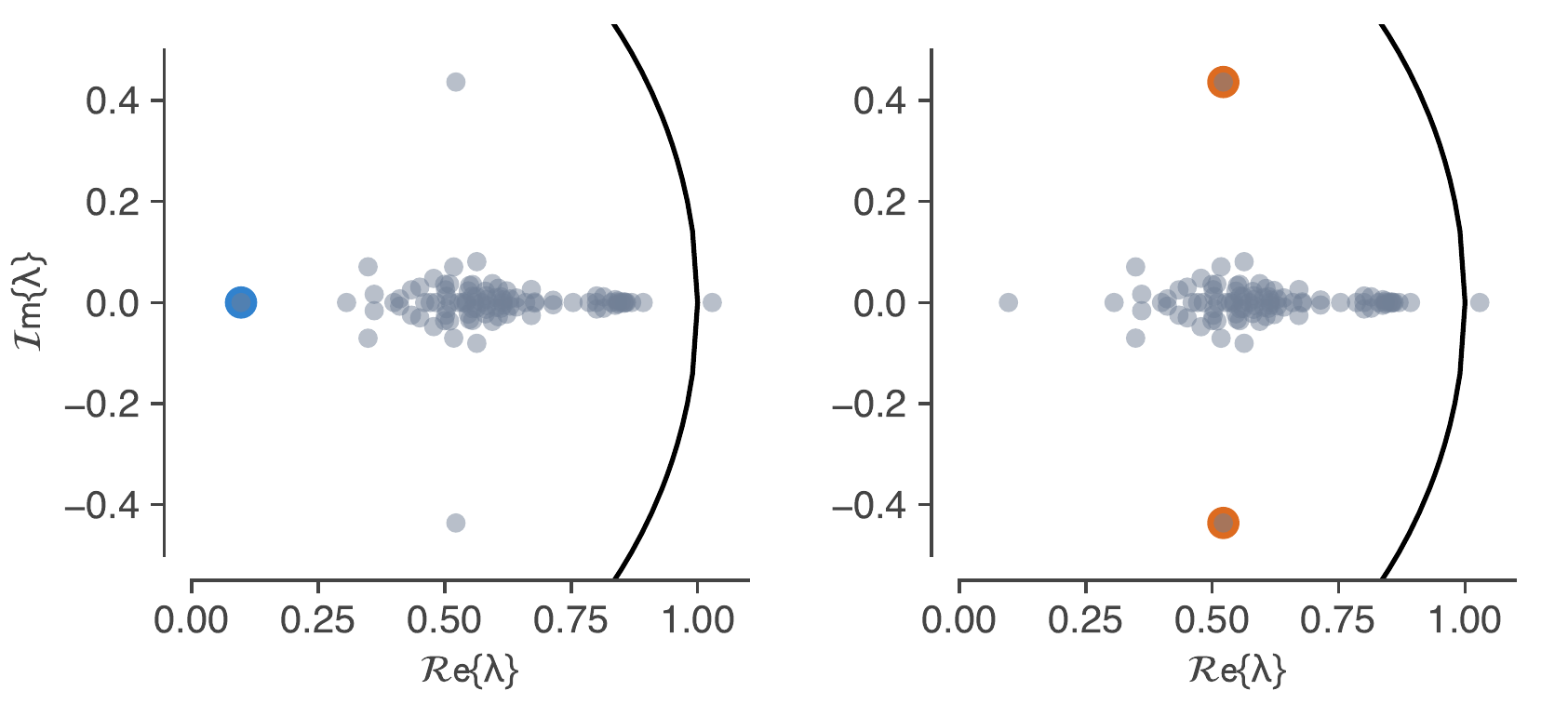}}
\caption{GRU trained on toy sentiment task. Shown are the eigenvalue plots for two fixed points along the line attractor. These fixed points are the ones closest to the $\bh^{\text{extremely}}$ and $\bh^{\text{not}}$ operating points, on the left and right panels, respectively. In both plots there is an integration mode near $(1,0)$.  In the left panel, the eigenvalue associated with the mode for the single-step decay dynamic for ``\emphasis{extremely}'' is shown in blue. In the right panel, the eigenvalues associated with the decay dynamic for the ``\negation{not}'' token are shown in orange.  The ``\emphasis{extremely}'' emphasis was defined for only the next word, while the ``\negation{not}'' negation was defined for four time steps.}
\label{fig:toy-eig-plots}
\end{center}
\vskip -0.2in
\end{figure}

We analyzed the trained networks using the methods developed by~\citet{maheswaranathan2019b}. An example state-space plot from a trained network is shown in Figure \ref{fig-synth-intro}. We verified that the network indeed has line attractor dynamics, as in ~\cite{maheswaranathan2019b}. More precisely, using fixed point optimization (see Supplemental Methods Section \ref{supp:fps}) we identified a 1D manifold of slow points as shown in Figure \ref{fig-synth-intro}. Each one of these slow points was approximately marginally stable, meaning in this case that a single mode of the linearized system was marginally stable. This can be seen in the eigenvalue plot in Supp. Figure \ref{fig:toy-eig-plots}. The eigenvalue associated with this mode is very close to $(1,0)$ in the complex plane.

\subsection{Analysis of state deflections caused by inputs after modifier tokens}
\begin{figure}[ht]
\vskip 0.2in
\begin{center}
\centerline{\includegraphics[width=3.4in]{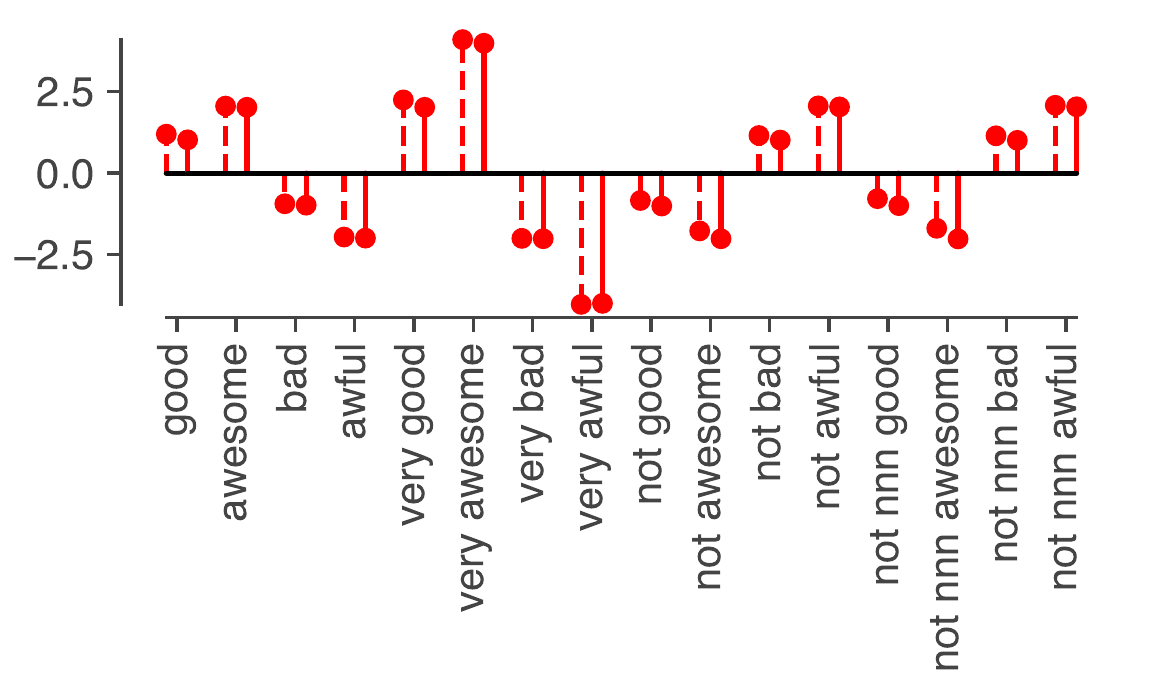}}
\caption{The effect of modifier tokens in the GRU trained on toy sentiment task. We computed $\Jinp\evalat{\bh^e, \bzero}
\; \bx$ as described in the text for both modifier inputs and non-modifier inputs. We then measured the readout values for all valence tokens and modifiers (red dashed lines). We also measured the readout value for the full nonlinear update from $\bh^e$ (red solid lines). The linear update on the readout is approximate to 91.8\% as measured by the mean absolute value of the ratio of the approximation over the full nonlinear update. Note that ``nnn'' means three noise words between ``not'' and the valence word.}
\label{fig:toy-inp-jac-approx}
\end{center}
\vskip -0.2in
\end{figure}

The first analysis we performed beyond those done in \cite{maheswaranathan2019b} was to examine whether effects of modifier tokens could be well-approximated by a carefully chosen linearization. We chose 
\begin{equation}
    \bh_t \approx F(\hmod, \bzero) + \Jinp\evalat{(\bh^{\text{mod}}, \bzero)} \bx_t, \label{eq:lin_inp_supp}
\end{equation}
with $\bh^{\text{mod}}$ denoting the state immediately after a modifier word enters the system. We measured the quality of equation (\ref{eq:lin_inp_supp}) by computing the output value 
\begin{equation}
    \bw^T\Jinp\evalat{(\bh^\te, \bzero)} \bx_t + b, \label{eq:lin_inp_ro}
\end{equation}
for all valence words when $\bh^e$ was set to be either 1) the state somewhere along the line attractor (denoted $\bh^*$), which measures normal integration dynamics, or 2) the state after a modifier word (denoted $\bh^{\text{mod}}$), which measures modified integration dynamics, so $\bh^e \in \{\bh^*, \bh^{\text{mod}}\}$. We show the results in Supp. Figure \ref{fig:toy-inp-jac-approx}. The results show the linear expansion is very good, yielding a 92\% accuracy in comparison to the full nonlinear update.

The quality of these expansion results now gives an abstract explanation for how the RNN implements context via modification words. A modifier enters the system and causes a deflection to $\bh^{\text{mod}}$. The computational purpose for this deflection is to enter a region of state space that modifies processing of subsequent word(s).  This can be measured by $\Jinp$, as in equation (\ref{eq:lin_inp_supp}) and to good approximation is essentially a linear update to $\bh^{\text{mod}}$, as measured by the effect on the readout.

\subsection{Analysis of state dynamics that participate in modifier computation}
Having established how modifier words cause contextual state deflections we became interested in how the RNN implemented the transient response to each modifier. In particular, as we defined the effect of ``extremely'' to last one token and the effect of ``not'' to last for four, we expected there to be different length transients induced by these modifiers.  To investigate this, we studied the linear system defined by 
\begin{equation}
    \Delta \bh_{t} \approx \Jrec\evalat{(\bh^*, \bx^*)} \Delta \bh_{t-1},  \label{eq:lin_rec_sup}
\end{equation}
where $\bx^*=\bzero$, and $\bh^*$ was a fixed point along the line attractor chosen because it was closest to $\bh^{\text{mod}}$, where  $\bh^{\text{mod}}=F(\bh_0, \bx^{\text{mod}})$ and $\bh_0$ is the system initial condition. Here $\bx^{\text{mod}}$ ranges over ``extremely'' and ``not''.

\begin{figure}[ht]
\vskip 0.2in
\begin{center}
\centerline{\includegraphics[width=0.7\columnwidth]{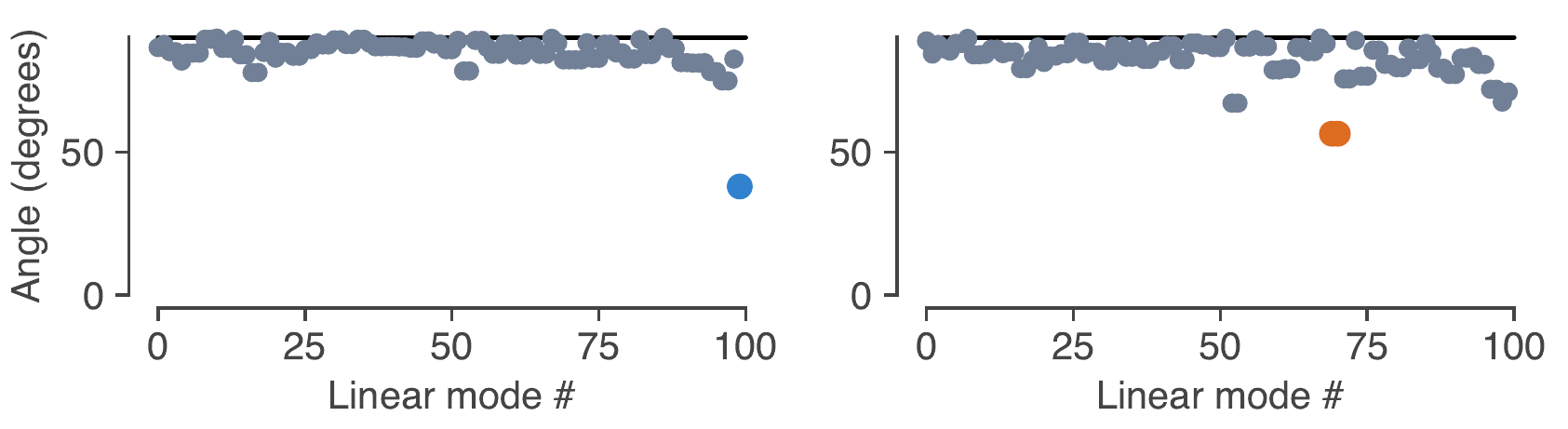}}
\caption{Subspace angles between state space deflections caused by modifier tokens and the left eigenvectors of the linearized systems around fixed points.  Left panel. The subspace angles (degrees) between $(\bh^\text{extremely}\!-\!\bh^*)$ and $\bell_a$ for the $a^{th}$ left eigenvector of the linear system around the fixed point closest to $\bh^{\text{extremely}}$. Right panel is the same as the left panel, except for analysis of the ``not'' modifier.}
\label{fig:toy-fixed-point-mode-angle}
\end{center}
\vskip -0.2in
\end{figure}

Examination of the eigenvalue spectra for linearizations around the fixed points along the line attractor (Supp. Figure \ref{fig:toy-eig-plots}) revealed three obvious modes far away from the bulk of the eigenvalue distribution. We wondered whether any of these three modes were responsible for implementing the transient dynamics associated with the two modifier words.  We reasoned that if these modes of were of any utility to modifier dynamics, then a projection of the state onto these modes should be apparent when a modifier word enters the RNN. Therefore, we computed the subspace angle between $(\bh^{\text{mod}}$ - $\bh^*)$ and $\bell_a$, the $a^{th}$ left eigenvector of the recurrent Jacobian from equation (\ref{eq:lin_rec_sup}).  The results are shown in Supp. Fig. \ref{fig:toy-fixed-point-mode-angle} and indeed show that the isolated extremely fast decaying mode is related to processing of ``extremely'' and that the oscillatory mode is related to processing the ``not'' token. It is the modes with smallest subspace angle that are highlighted with color in Supp. Fig. \ref{fig:toy-eig-plots}.

\begin{figure}[ht]
\vskip 0.2in
\begin{center}
\centerline{\includegraphics[width=4.25in]{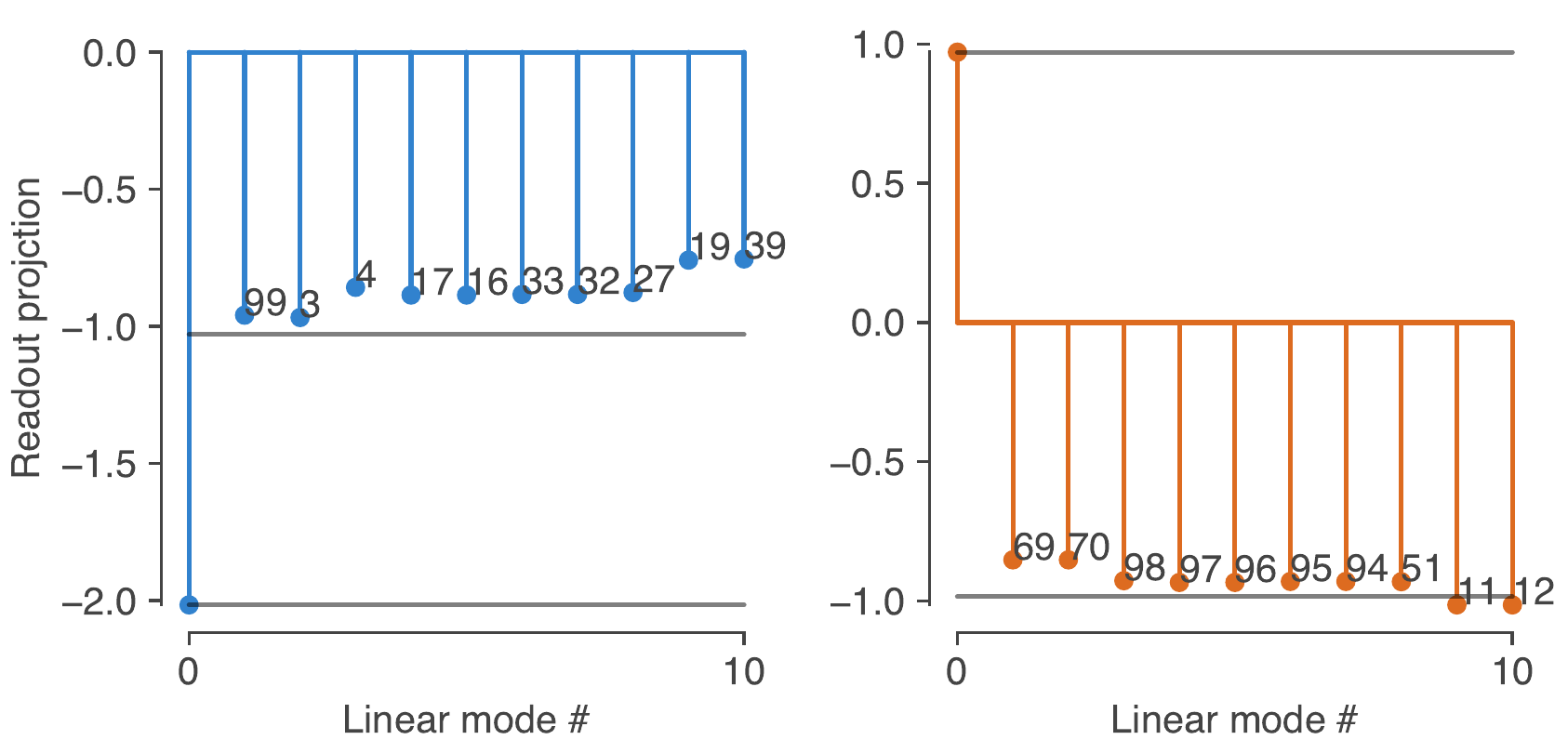}}
\caption{Removing the components of $\bh^{\text{mod}}$ that project onto select left eigenvectors of the linearized systems around fixed points completely destroys the effect of the modifier input. See text for methodological details. Left) The result of removing $\bell_{99}$, a very fast decaying mode, completely destroyed the effect of the ``extremely'' modifier token on the valence word ``bad''.  Right) removing the modes $\bell_{69}$ and $\bell_{70}$ completely destroyed the effect of the ``not'' modifier token on the valence word ``bad''.  In this case there were three noise words used, e.g. ``not the the the bad''.}
\label{fig:toy-mode-removal}
\end{center}
\vskip -0.2in
\end{figure}

Finally, we reasoned that if these modes of the linearized systems were important for processing the effect of modifier inputs, then removing the projection onto those modes should remove the effect of the modifier.  We therefore rank ordered the left eigenvectors by how much their individual removal perturbed the effect of the modifier. Specifically, we defined the state deflection of a valence word after a modifier word as  $\bh^{\text{valence}}=F(\bh^{\text{mod}}, \bx^{\text{valence}})$, then we measured error in $\bw^T \bh^{\text{valence}} + \mathbf{b}$ as the components of $\bh^{\text{mod}}$ that projected onto $\bell_a$ were removed, for all $a$. We then computed the cumulative sum of these effects, starting with the removal of the mode that was most important and then removing both the first and second most important modes, etc.\footnote{For complex conjugate pairs, we always removed the pair when one vector was slated to be removed.} The results are shown in Supp. Fig. \ref{fig:toy-mode-removal} and confirm that removing even the single mode with largest projection into $(\bh^{\text{mod}}\!-\!\bh^*)$ essentially completely destroys the modification effect.

\section{Fixed point finding methodology} \label{supp:fps}
We treat any recurrent neural network (RNN) update as a function $F$, that updates a hidden state $\bh$ in response to some input: $\bh_{t} = F(\bh_{t-1}, \bx_t)$. This defines a discrete time dynamical system. \textit{Fixed points} of these dynamics are given by points $\bh^*$ where applying the function $F$ does not change the state: $\bh^* = F(\bh^*, \bx)$, for some input $\bx$. Here, we focus on fixed points in the absence of input ($\bx = \bzero$).

To computationally identify fixed points of the recurrent dynamics, we numerically solve the following optimization problem~\cite{Sussillo2013,Golub2018}:
\begin{equation}
    \min_{\bh} \frac{1}{2} \| \bh - F(\bh, \bzero) \|_2^2. \label{eq:fp_finding}
\end{equation}
In general, the function minimized in equation (\ref{eq:fp_finding}) has many local minima, corresponding to different slow points or fixed points of $F$. We are interested in finding all slow points, regardless of whether they are local minima, such that $|\bh^*-F(\bh^*, \bzero)|$ is significantly smaller than $|\Jrec\evalat{(\bh^*, \bzero)} \Delta \bh_{t-1}|$ in $\bh_{t} \approx F(\bh^*, \bzero) + \Jrec\evalat{(\bh^*, \bzero)} \Delta \bh_{t-1}$, thereby enabling high-quality linear approximations to the dynamics. We find these by running the optimization problem above starting from a large set of initial points, taken from the randomly selected RNN hidden states visited during test examples~\cite{maheswaranathan2019b}. We initialized the fixed point finding routine using 10,000 hidden states of the RNNs while performing sentiment classification.

\section{Additional architectures}\label{supp:lstm}

\begin{figure}[ht]
    \begin{center}
    \centerline{\includegraphics[width=0.8\columnwidth]{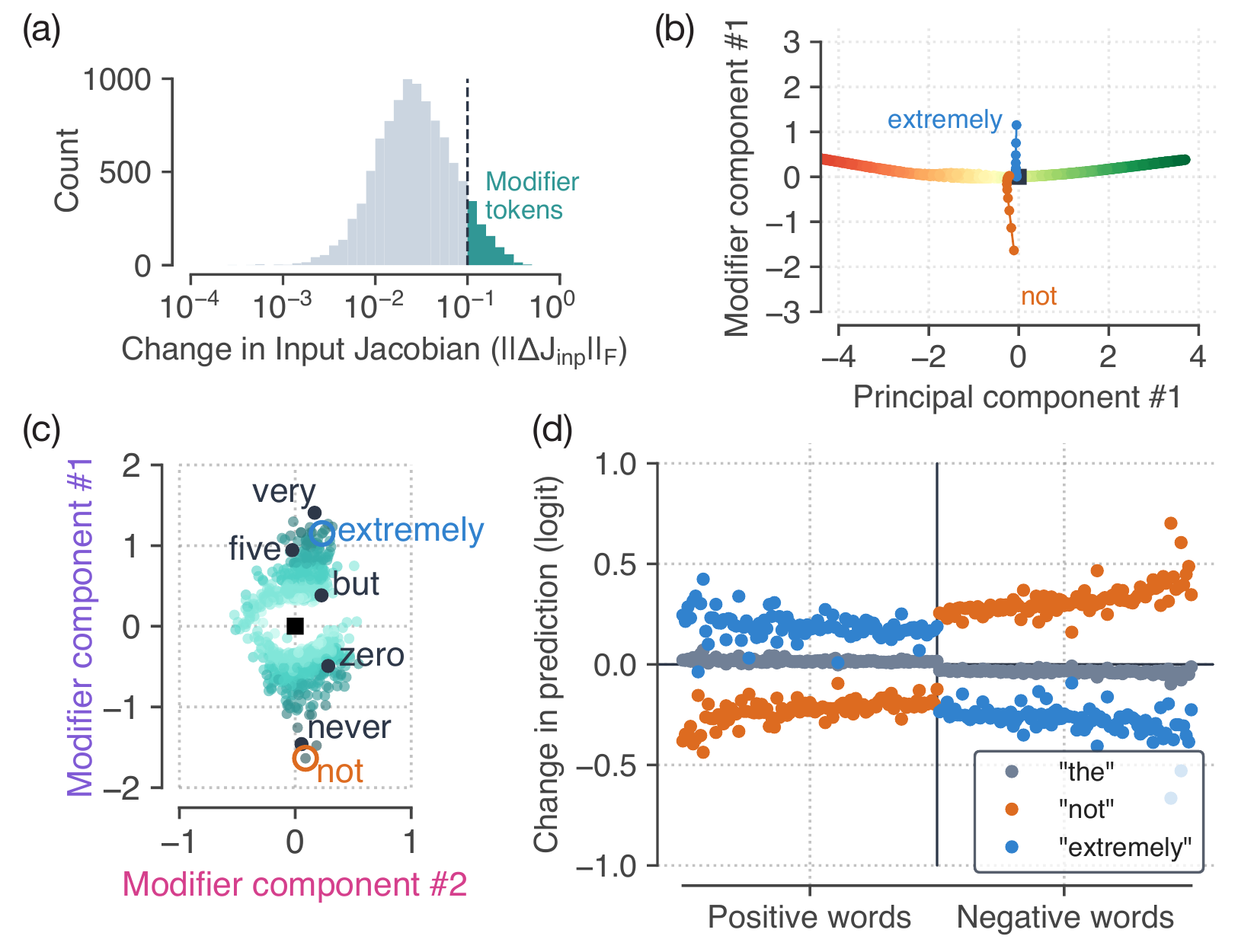}}
    \caption{Analysis of modifier effects for an LSTM. This figure reproduces the analyses for an LSTM that are shown in the main paper for a GRU. It shows that LSTMs implement contextual effects of modifier words in a way very similar to the GRU. (a) Histogram of modifiers using data driven approach, as in Fig.~\ref{fig:dij}. (b) Impulse response to modifier words, as in Fig.~\ref{fig:impulse}a. (c) Modifier subspace showing arrangement of various modifier words, as in Fig.~\ref{fig:subspace}. (d) Modifier barcodes for ``\negation{not}'', ``the'', and ``\emphasis{extremely}'' as in Fig.~\ref{fig:barcodes}. }
    \label{fig:supp-lstm}
    \end{center}
    \vskip -0.2in
\end{figure}

In the main text, we provided a detailed analysis of a particular RNN, a gated recurrent unit (GRU)~\cite{Cho2014}. We were interested in whether our results were sensitive to this particular RNN architecture. To explore this, we trained an analyzed additional RNN architectures on the same task. Supp. Fig. \ref{fig:supp-lstm} shows the results of applying our analyses to a trained LSTM~\cite{Hochreiter1997}. We find a remarkable degree of similarity between the two networks (compare Fig.~\ref{fig:dij}, Fig.~\ref{fig:impulse}a, Fig.~\ref{fig:subspace}, and Fig.~\ref{fig:barcodes} from the main text to Supp. Fig.~\ref{fig:supp-lstm}a-d, respectively).

We see similar results for the other gated RNN architecture we trained, the Update Gate RNN (UGRNN;~\cite{Collins2016}). However, for the Vanilla RNN, we find that it does \textit{not} have a clear modifier subspace, and is incapable of correctly processing contextual effects (Fig.~\ref{fig:supp-arch}). We suspect this is due to difficulties in training vanilla RNNs, rather than due to reduced modeling capacity, as the mechanisms proposed in this paper can in theory be implemented using Vanilla RNNs.

The degree of similarity in the mechanisms learned by gated RNNs for contextual processing suggest that these mechanisms may be universal computational motifs~\cite{maheswaranathan2019a}.

\section{Additional barcodes}\label{supp:barcodes}

\begin{figure}[ht]
    \begin{center}
    \centerline{\includegraphics[width=\columnwidth]{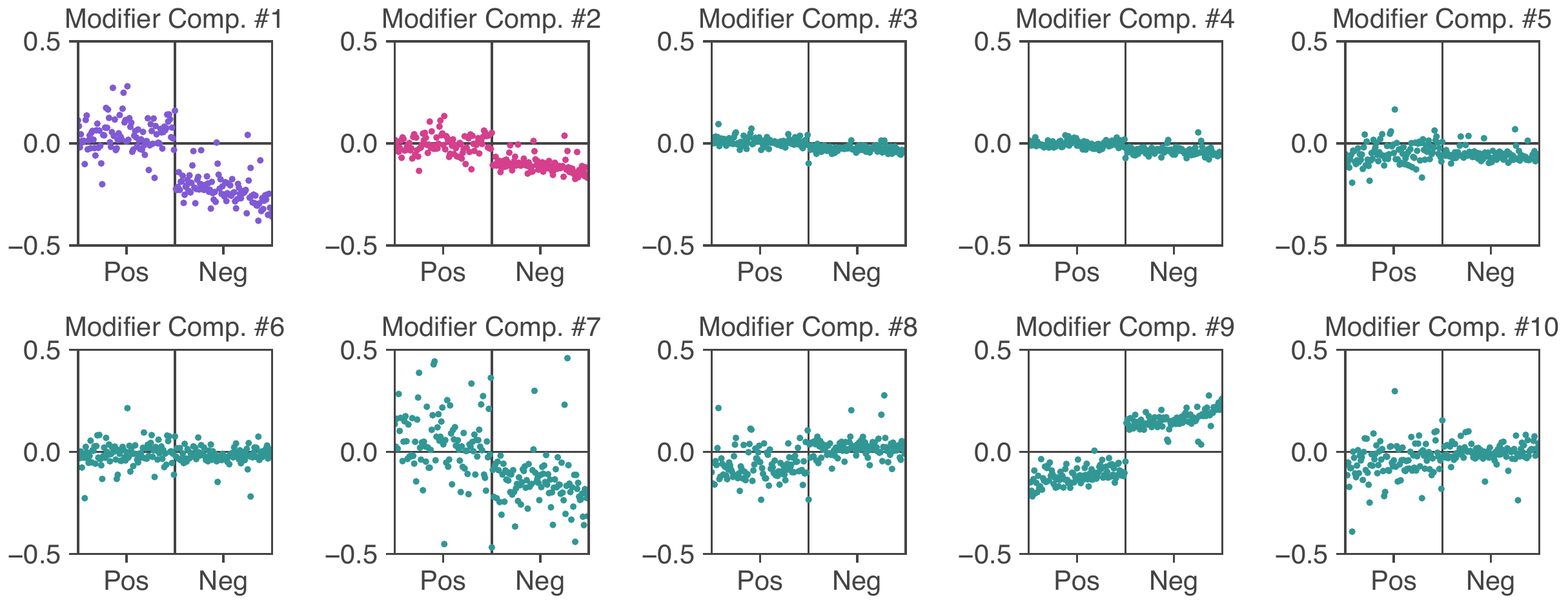}}
    \caption{Additional modifier barcodes. For every modifier component, we compute a barcode to summarize the effect of that particular modifier dimension on inputs. The barcode summarizes the change in sensitivity to particular salient inputs (here, the top 100 positive and top 100 negative words). Barcode values of zero indicate that the sensitivity to those words is not affected (that is, there are no contextual effects). Each panel shows the barcode corresponding to a given modifier component. We additionally highlight the first (purple) and second (pink) components using different colors.}
    \label{fig:additional-barcodes}
    \end{center}
    \vskip -0.2in
\end{figure}

The barcodes presented in Figure \ref{fig:bilinear_barcodes} show the effects of modifiers along the top two modifier components (colored in purple and pink, respectively). Below (Fig.~\ref{fig:additional-barcodes}), we present additional barcodes for the top 10 modifier components. Note that nearly all of the variance in the modifier subspace is captured by just a few (2-3) components. Perturbation experiments also suggest that contributions from modifier components outside the top three to overall accuracy are minor~(Supp. Fig.~\ref{fig:perturbation}, \S\ref{supp:perturbations}). However, these additional barcodes do contain interesting structure. In particular, components seven and nine may have subtle contributions to processing of modifiers.

\section{Transient EOD effects are implemented using two modes}
\label{supp:eod}

\begin{figure}[ht]
    \begin{center}
        \centerline{\includegraphics[width=\columnwidth]{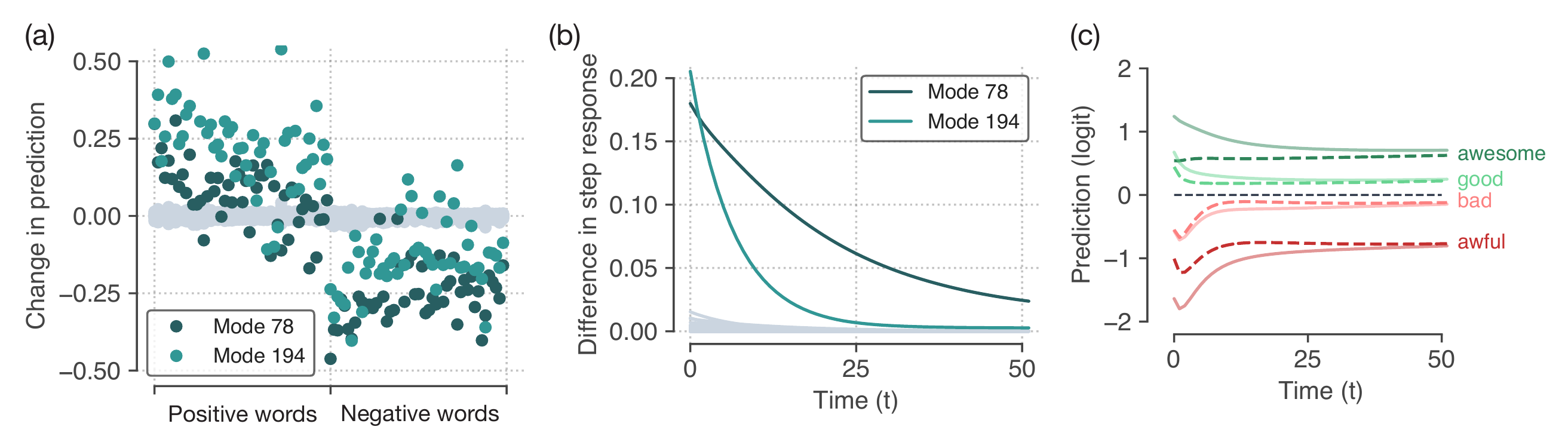}}
        \caption{An analysis of transient linear dynamics that contribute to the end of document contextual effects. (a) We analyzed the linear modes of the recurrent Jacobian around a typical fixed point on the line attractor. In particular, we studied the projection of the linear modes onto the readout as a function of eigenvalue decay.  We found two modes, \#78, and \#194 (dark green and light green respectively) that had substantial projections onto the readout and correspondingly large effects on the transient valence of positive and negative words (gray shows the change in prediction for the other modes). (b) The average absolute change in transient valence to valence words achieved by modes \#78 and \#194. (c) The transient valence of ``awesome'', ``good'', ``bad'', ``awful'' are  reproduced from Figure 9 (solid lines), while the effect of removing the projection of the readout onto transient modes \#78 and \#194 are also shown (dashed lines).}
        \label{fig:supp_transients}
    \end{center}
\end{figure}

We observed contextual processing at the end of a document, implemented using transient decay dynamics. That is, valence words would initially induce a large projection onto the readout, but this would decay to a steady-state value (Fig.\ref{fig:boseos}b). The steady-state effect occurs due to integration along the line attractor~\cite{maheswaranathan2019b}.

For the transient effects, we studied modes outside of the integration modes (those associated with slow eigenvalues) and modifier dynamics (those associated with very fast modes on the timescale of tens of tokens). In particular, we found two additional modes which were strongly activated by valence words, mode 78 (with a corresponding timescale of 19.57 tokens) and mode 194 (with a timescale of 6.63 tokens).

To quantify the transient suppression for these two modes, we computed the (instantaneous) change in prediction when we applied the linearized RNN with individual modes (eigenmodes) removed. This allows us to examine how important individual modes are to the instantaneous processing of valence tokens. Removing modes 78 and 194 resulted in strong changes in the valence associated with positive and negative words (Supp. Fig.~\ref{fig:supp_transients}a). To verify the timescale of these transient effects, we computed the average change in prediction across a set of 100 valence (50 positive and 50 negative) words over time. This allows us to investigate how individual eigenmodes contribute to the overall transient (Supp. Fig.~\ref{fig:supp_transients}b). Finally, we performed a perturbation experiment to examine the effect of these two modes in particular. We probed the system's step response to valence tokens (``awesome'', ``good'', ``bad'', and ``'awful''), and either ran the full system or projected the hidden state out of subspace defined by modes 78 and 194 (Fig.~\ref{fig:supp_transients}c). This perturbation reduced or eliminated the transient effects.

Taken together, these results suggest that the RNN implements end of document contextual processing using two additional modes of the recurrent dynamics. These modes induce a transient that decays with timescales of around 7 and 20 tokens. Therefore, the last 20 or so tokens in a given document will be emphasized relative to those in the middle of a document.

\section{Perturbations}
\label{supp:perturbations}

\begin{figure}[ht]
    \vskip 0.2in
    \begin{center}
    \centerline{\includegraphics[width=\columnwidth]{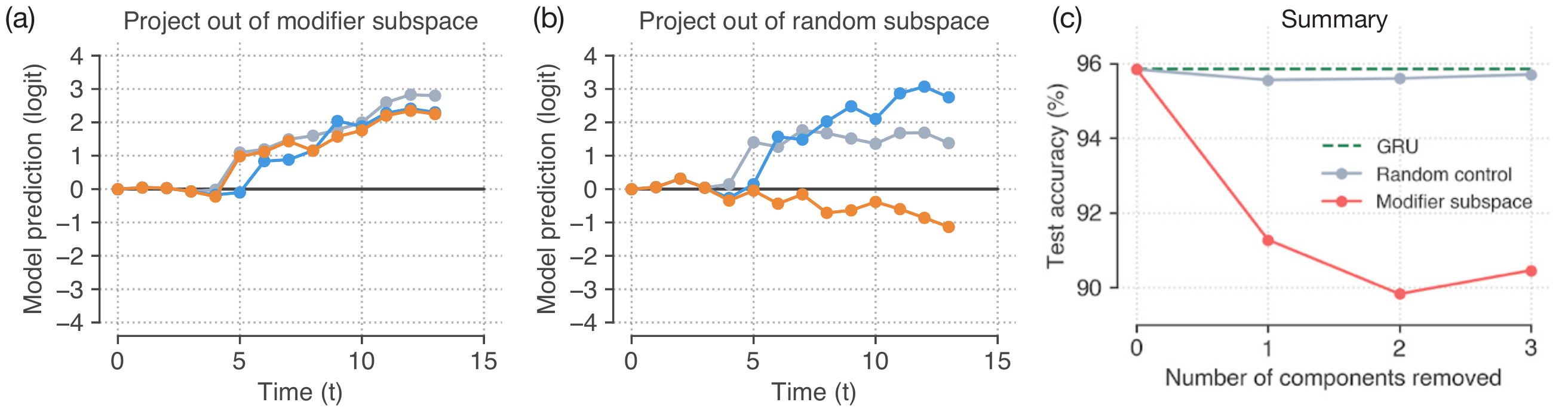}}
    \caption{Perturbation experiment demonstrates that the modifier subspace is necessary for contextual processing. (a) Perturbation where a trained network is probed with three example sentences, ``This movie is awesome. I like it.'' (gray), ``This movie is \negation{not} awesome, I \negation{don't} like it.'' (orange), and ``This movie is \emphasis{extremely} awesome, I \emphasis{definitely} like it.'' (blue), and where the hidden state is projected \textit{out} of the modifier subspace (here, the modifier subspace is three dimensional) at each iteration. This perturbation collapses the trajectories, indicating that the system is no longer capable of processing modifier tokens. Moreover, the effect of this perturbation is to selectively remove contextual effects, but does not affect the baseline integration. (b) Random control where we project the hidden state out of a \textit{random} three dimensional subspace. We see that this has no effect on the network activity (compare with Fig.~\ref{fig:recap}b). (c) Summary across all test examples. Performance of original network (dashed green line) compared with perturbations where we project the hidden state at each timestep or iteration when applying the RNN. Projecting out of a random three dimensional subspace (solid gray line) has a negligible effect on the accuracy, whereas projecting out of the modifier subspace (solid red line) has a significant effect, equivalent to shuffling input tokens (Fig.~\ref{fig:supp-arch}a).}
    \label{fig:perturbation}
    \end{center}
    \vskip -0.2in
\end{figure}

Here, we perform perturbation experiments to eliminate particular mechanisms in the RNN. This demonstrates that these mechanisms are necessary for particular function. To test whether the modifier subspace is necessary for contextual processing, we probed the RNN with the same examples from Fig.~\ref{fig:recap}a, but at every step we project the RNN hidden state \textit{out} of either the modifier subspace (Supp. Fig. \ref{fig:perturbation}a) or a random control subspace (Supp. Fig. \ref{fig:perturbation}b). The modifier subspace perturbation selectively removes the network's ability to process contextual inputs. Finally, we ran the same perturbation for all of the examples in the test set. We found a negligible effect on accuracy when projecting out of a random subspace, but a significant reduction in accuracy when projecting out of the modifier subspace (Supp. Fig.~\ref{fig:perturbation}c).

\section{Augmented baseline models}
\label{supp:baselines}

Below, we provide full definitions for all of the augmented baseline models tested in \S\ref{baselines}.

\textbf{Bag-of-words model} ($W+1$ parameters): Each word in the document is associated with a scalar weight. These weights are then summed along with a bias.
\begin{equation}
    \sum_{t=1}^T \beta[t] + b
\end{equation}

\textbf{Convolution of Modifier Words (CoMW)} ($W + 1 + 2 M$ parameters): Modifier words scale the same weights used to estimate the valence of each word without modification. The modifier scaling decays exponentially with time.
\begin{align}
 \sum_{t=1}^T \left(\beta[t] + \sum_{m=1}^{M} \left(f^m \ast \mu^m\right)[t] \; \beta[t]\right) + b
\end{align}

\textbf{Convolution of BOD and EOD Tokens} ($W + 1 + 4$ parameters):
We augmented each example with Beginning of Document (BOD) token and End of Document (EOD) tokens. For the EOD token, the exponential filter was flipped acausally.
\begin{equation}
 \sum_{t=1}^T \left(\beta[t] + \sum_{m \in \left[\mbor, \meor\right]} \left(f^m \ast \mu^m\right)[t] \; \beta[t]\right) + b
\end{equation}
 
\textbf{Convolution of Modifier Words + $\boldsymbol{\beta}_\text{mod}$} ($W + 1 + 2 M + P W$ parameters): Modifier words exponentially scale additional learned weight vectors used exclusively to estimate modification. 
\begin{equation}
    \sum_{t=1}^T \left(\beta[t] + \sum_{p=1}^{P}\sum_{m=1}^M \left(f^m \ast \mu^m\right)[t] \; \beta^p_{\text{mod}}[t] \right) + b 
\end{equation}

\textbf{Convolution of BOD and EOD Tokens + $\boldsymbol{\beta}_\text{BOD} + \boldsymbol{\beta}_\text{EOD}$} ($W + 1 + 4 + 2 W$ parameters):
Similar to Convolution of BOD and EOD Tokens, except that we allow the learning of two separate $\mathbf{\beta}_{\text{mod}}$ weights to estimate the effects of both the beginning and end of the review. 
\begin{equation}
    \sum_{t=1}^T \left(\beta[t] + \sum_{m\in[\mbor, \meor]} \left(f^m \ast \mu^m\right)[t] \; \beta^m_{\text{mod}}[t]\right) + b
\end{equation}

\textbf{Convolution of Modifier Words + $\boldsymbol{\beta_\text{mod}} + \boldsymbol{\beta_\text{BOD}} + \boldsymbol{\beta_\text{EOD}}$} ($W + 1 + 4 + 2W + 2 M + P W$ parameters):  Combines the most powerful model versions from above to learn both modifier word effects as well as contextual effects at the beginning and end of the review.
\begin{equation}
    \sum_{t=1}^T \left( \beta[t] + \sum_{m \in [\mbor, \meor]} \left(f^m \ast \mu^m\right)[t] \; \beta^m_{\text{mod}}[t] \:+ \sum_{p=1}^{P}\sum_{m=1}^M \left(f^m \ast \mu^m\right)[t] \; \beta^p_{\text{mod}}[t]\right) + b
\end{equation}